\documentclass[12pt]{article}
\usepackage[left=2.5cm,
right=2.5cm,
top=2.3cm,
bottom=2.3cm,
headheight=20pt,
headsep=10pt,
footskip=25pt,
letterpaper]{geometry}
\usepackage[utf8]{inputenc}
\usepackage[T1]{fontenc}
\usepackage[english]{babel}
\usepackage{xcolor}
\usepackage{url}
\usepackage{amsfonts}
\usepackage{nicefrac}
\usepackage{microtype}
\usepackage{amsmath}
\usepackage{amssymb}
\usepackage{mathtools}
\usepackage{subcaption}
\usepackage{ragged2e}
\usepackage{stackengine}
\usepackage{etoolbox}
\usepackage{xspace}
\usepackage{xpatch}
\usepackage{cuted}
\usepackage{enumerate}
\usepackage{xstring}
\usepackage{natbib}
\usepackage{setspace}
\usepackage{makecell}
\usepackage{graphicx}
\usepackage{changepage}
\usepackage{enumitem}
\usepackage{eqparbox}
\usepackage{wrapfig}
\usepackage[hang,flushmargin,bottom,symbol]{footmisc}
\usepackage[useregional=numeric]{datetime2}
\usepackage{pifont}
\usepackage{listings}
\usepackage{environ}
\usepackage{tabularray}
\usepackage{titlesec}
\usepackage{titletoc}
\usepackage{afterpage}
\usepackage{fancyhdr}
\usepackage{trimclip}
\usepackage[colorlinks]{hyperref}
\usepackage[capitalise,noabbrev,nameinlink]{cleveref}
\usepackage{lineno}
\usepackage{float}

\fancypagestyle{first}{\fancyfoot[R]{\small\thepage}}

\setlist[itemize]{leftmargin=1em,itemsep=0ex,topsep=0ex}
\titlespacing*{\paragraph}{0pt}{0ex plus .1ex}{1ex}
\titlespacing*{\section}{0ex}{2.3ex plus .3ex minus .0ex}{.6ex plus .3ex minus .2ex}
\titlespacing*{\subsection}{0ex}{1.5ex plus .3ex minus .5ex}{.4ex plus .2ex minus .1ex}
\titlespacing*{\subsubsection}{0ex}{1.2ex plus .3ex minus .3ex}{.3ex plus .2ex minus .2ex}

\xapptocmd\normalsize{%
\abovedisplayskip=.8em plus .2em minus .2em
\belowdisplayskip=.6em plus .1em minus .1em
\abovedisplayshortskip=.8em plus .2em minus .2em
\belowdisplayshortskip=.6em plus .1em minus .1em
}{}{}

\newcommand{\itempar}[1]{\item\textbf{#1}\quad}
\newcommand{\padspace}{\hspace{3.5em}}

\setcitestyle{numbers}
\renewcommand{\cite}[1]{\citep{#1}}

\creflabelformat{equation}{#2\textup{#1}#3}
\crefname{algocf}{Algorithm}{Algorithms}
\Crefname{algocf}{Algorithm}{Algorithms}

\definecolor{mydarkblue}{rgb}{0.0,0.15,0.7}
\hypersetup{%
colorlinks=true,
linkcolor=mydarkblue,
citecolor=mydarkblue,
filecolor=mydarkblue,
urlcolor=mydarkblue}

\graphicspath{{figures/}}

\makeatletter
\def\hlinewd#1{%
\noalign{\ifnum0=`}\fi\hrule \@height #1 \futurelet
\reserved@a\@xhline}
\makeatother
\UseTblrLibrary{booktabs}
\NewColumnType{L}[1]{Q[l,#1]}
\NewColumnType{C}[1]{Q[c,#1]}
\NewColumnType{R}[1]{Q[r,#1]}
\renewcommand{\o}{\hphantom{0}}
\NewEnviron{mytabular}[1]{%
\setlength\heavyrulewidth{1.2pt}
\setlength\lightrulewidth{.5pt}
\setlength\aboverulesep{.4ex}%
\setlength\belowrulesep{.8ex}%
\begin{booktabs}[expand=\BODY]{#1}
\BODY
\end{booktabs}
}

\makeatletter
\DeclareDocumentCommand\todo{g}{%
\def\@message{\IfNoValueTF{#1}{TODO}{TODO: #1}}
\textbf{\textcolor[HTML]{FF8811}{\@message}}
\@latex@warning{\@message}{}{}}
\makeatother

\newcommand{\blap}[1]{\vbox to 0pt{\hbox{#1}\vss}}

\renewcommand\d{\mathop{}\!\textnormal{\slshape d}}

\makeatletter
\newcommand{\removeParBefore}{\ifvmode\vspace*{-\baselineskip}\setlength{\parskip}{0ex}\fi}
\newcommand{\removeParAfter}{\@ifnextchar\par\@gobble\relax}
\newcommand{\eq}{\begingroup\removeParBefore\endlinechar=32 \eqinner}
\newcommand{\eqinner}[2][aligned]{\endlinechar=32%
\begin{gather}\begin{#1}#2\end{#1}\end{gather}\endgroup\removeParAfter}
\makeatother

\DeclareDocumentCommand{\p}{ D<>{p} D<>{} r() }{
\def\content{#3}\patchcmd{\content}{|}{\;#2\vert\;}{}{}
\ensuremath{#1 #2(\content #2)}}

\DeclareDocumentCommand{\P}{ D<>{P} D<>{\big} r() }{
\def\content{#3}\patchcmd{\content}{|}{\;#2\vert\;}{}{}
\ensuremath{\operatorname{#1}#2(\content #2)}}

\DeclareDocumentCommand{\E}{ D<>{E} E{_}{{}} D<>{\big} r[] }{
\def\content{#4}\patchcmd{\content}{|}{\;#3\vert\;}{}{}
\ensuremath{\operatorname{#1}_{#2}#3[\content #3]}}

\DeclareDocumentCommand{\D}{ D<>{D} D<>{\big} r[] }{
\def\content{#3}\patchcmd{\content}{||}{\;#2\|\;}{}{}
\ensuremath{\operatorname{#1}\!#2[\content #2]}}

\NewDocumentCommand{\Nor}{ r() }{\P<Normal>](#1)}
\NewDocumentCommand{\Cat}{ r() }{\P<Cat>](#1)}
\NewDocumentCommand{\Bin}{ r() }{\P<Bin>](#1)}
\NewDocumentCommand{\Bet}{ r() }{\P<Beta>](#1)}
\NewDocumentCommand{\Ber}{ r() }{\P<Bernoulli>(#1)}
\NewDocumentCommand{\Dir}{ r() }{\P<Dir>(#1)}

\DeclareDocumentCommand{\KL}{ D<>{\big} r[] }{\D<KL><#1>[#2]}
\DeclareDocumentCommand{\H}{ D<>{\big} r[] }{\E<H><#1>[#2]}
\DeclareDocumentCommand{\I}{ D<>{\big} r[] }{\E<I><#1>[#2]}

\DeclareDocumentCommand{\lnpp}{ D<>{} r() }{
\ensuremath{\p<\ln p_\phi><#1>(#2)}}
\DeclareDocumentCommand{\pp}{ D<>{} r() }{
\ensuremath{\p<p_\phi><#1>(#2)}}
\DeclareDocumentCommand{\qp}{ D<>{} r() }{
\ensuremath{\p<q_\phi><#1>(#2)}}
\DeclareDocumentCommand{\SymLogNormal}{ D<>{} r() }{
\ensuremath{\p<\operatorname{SymLogNormal}><#1>(#2)}}
\newcommand{\sign}{\operatorname{sign}}

\newcommand{\eps}{\epsilon}

\newcommand{\symlog}{\ensuremath{\operatorname{symlog}}}
\newcommand{\symexp}{\ensuremath{\operatorname{symexp}}}
\newcommand{\twohot}{\ensuremath{\operatorname{twohot}}}
\newcommand{\sg}{\ensuremath{\operatorname{sg}}}

\newcommand{\EMA}{\operatorname{EMA}}
\newcommand{\Per}{\operatorname{Per}}

\title{\vspace*{-2.5ex}\bfseries Mastering Diverse Domains through World Models}

\date{}

\author{
Danijar Hafner\rlap{,}\textsuperscript{12}\enskip
Jurgis Pasukonis\rlap{,}\textsuperscript{1}\enskip
Jimmy Ba\rlap{,}\textsuperscript{2}\enskip
Timothy Lillicrap\textsuperscript{1}}

\begin{document}
\pagestyle{fancy}

\vspace{-4ex}
\maketitle
\thispagestyle{first}

\enlargethispage{1.5\baselineskip}
{\renewcommand\thefootnote{}\footnote{
\textsuperscript{1}Google DeepMind.\enskip
\textsuperscript{2}University of Toronto.\enskip
Correspondence: mail@danijar.com}}
\addtocounter{footnote}{-1}

\vspace{-6ex}
\begin{center}\bfseries
\vspace{-.5ex}
Abstract
\vspace{-.2ex}
\end{center}
\begin{adjustwidth}{0.95cm}{0.95cm}
\begin{hyphenrules}{nohyphenation}
\ignorespaces
Developing a general algorithm that learns to solve tasks across a wide range of applications has been a fundamental challenge in artificial intelligence.
Although current reinforcement learning algorithms can be readily applied to tasks similar to what they have been developed for, configuring them for new application domains requires significant human expertise and experimentation.
We present DreamerV3, a general algorithm that outperforms specialized methods across over 150 diverse tasks, with a single configuration.
Dreamer learns a model of the environment and improves its behavior by imagining future scenarios.
Robustness techniques based on normalization, balancing, and transformations enable stable learning across domains.
Applied out of the box, Dreamer is the first algorithm to collect diamonds in Minecraft from scratch without human data or curricula.
This achievement has been posed as a significant challenge in artificial intelligence that requires exploring farsighted strategies from pixels and sparse rewards in an open world.
Our work allows solving challenging control problems without extensive experimentation, making reinforcement learning broadly applicable.

\end{hyphenrules}
\end{adjustwidth}
\vspace{2ex}

\vfill
\begin{figure}[h]
\centering
\begin{adjustwidth}{0em}{-0.7em}
\includegraphics[width=\linewidth]{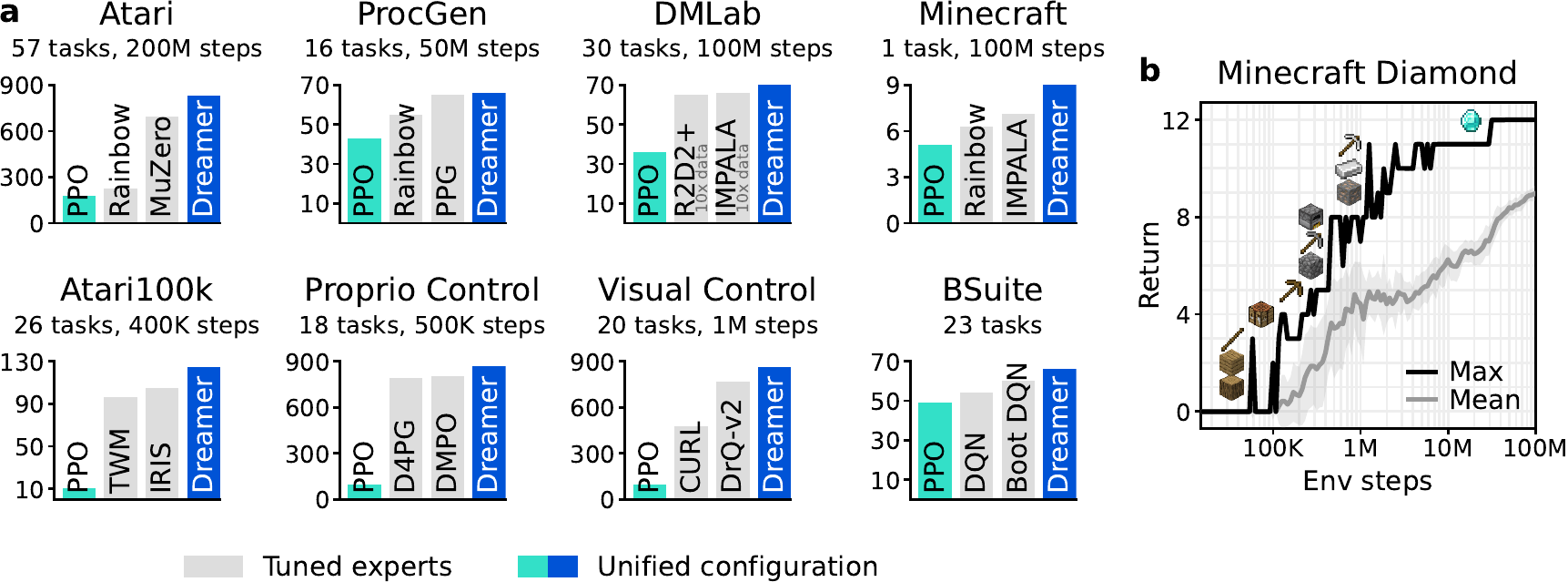}
\end{adjustwidth}
\vspace*{-0.5ex}%
\caption{Benchmark summary. \textbf{a}, Using fixed hyperparameters across all domains, Dreamer outperforms tuned expert algorithms across a wide range of benchmarks and data budgets. Dreamer also substantially outperforms a high-quality implementation of the widely applicable PPO algorithm. \textbf{b}, Applied out of the box, Dreamer learns to obtain diamonds in the popular video game Minecraft from scratch given sparse rewards, a long-standing challenge in artificial intelligence for which previous approaches required human data or domain-specific heuristics.}
\label{fig:bars}
\vspace*{-1ex}%
\end{figure}
\vfill
\clearpage

\begin{figure}[t]
\vspace*{-2ex}
\centering
\begin{subfigure}{.18\textwidth}
\includegraphics[width=\linewidth]{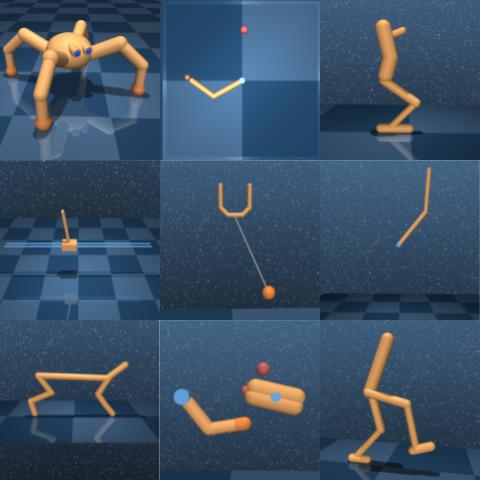}
\caption{Control Suite}
\end{subfigure}\hfill
\begin{subfigure}{.18\textwidth}
\includegraphics[width=\linewidth]{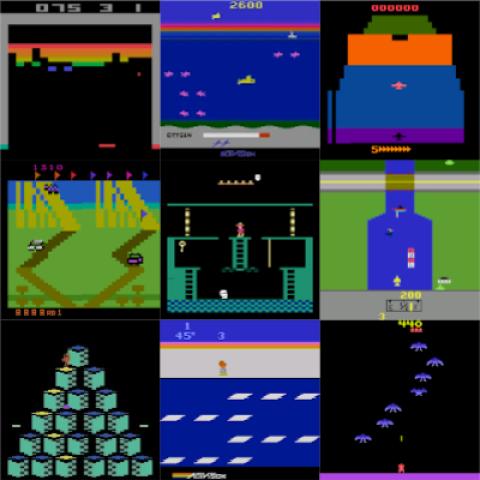}
\caption{Atari}
\end{subfigure}\hfill
\begin{subfigure}{.18\textwidth}
\includegraphics[width=\linewidth]{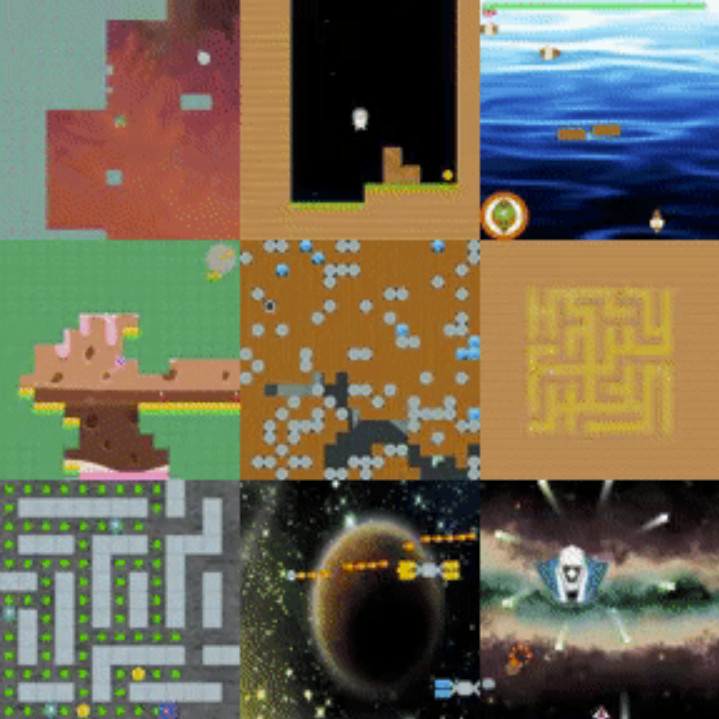}
\caption{ProcGen}
\end{subfigure}\hfill
\begin{subfigure}{.18\textwidth}
\includegraphics[width=\linewidth]{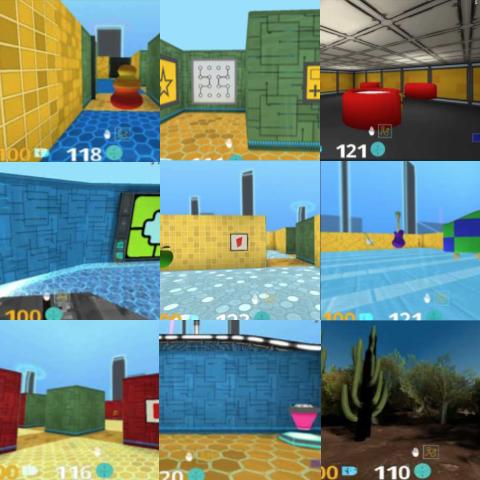}
\caption{DMLab}
\end{subfigure}\hfill
\begin{subfigure}{.18\textwidth}
\includegraphics[width=\linewidth]{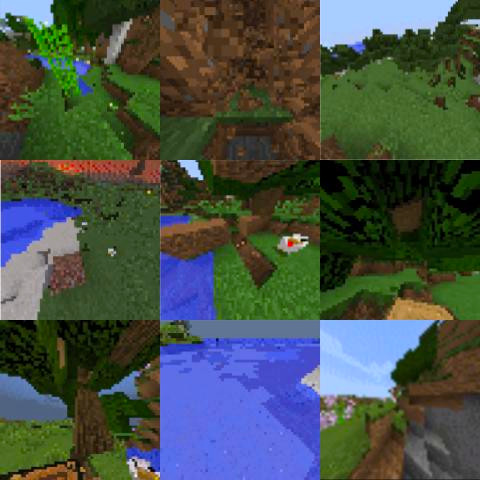}
\caption{Minecraft}
\end{subfigure}
\caption{Diverse visual domains used in the experiments. Dreamer succeeds across these domains, ranging from robot locomotion and manipulation tasks over Atari games, procedurally generated ProcGen levels, and DMLab tasks, that require spatial and temporal reasoning, to the complex and infinite world of Minecraft. We also evaluate Dreamer on non-visual domains.}
\label{fig:tasks}
\end{figure}

\enlargethispage{2ex}

\section*{Introduction}
\label{sec:intro}

Reinforcement learning has enabled computers to solve tasks through interaction, such as surpassing humans in the games of Go and Dota \citep{silver2016alphago,openai2018dota}.
It is also a key component for improving large language models beyond what is demonstrated in their pretraining data\citep{ouyang2022rlhf,le2022coderl}.
While PPO \citep{schulman2017ppo} has become a standard algorithm in the field of reinforcement learning, more specialized algorithms are often employed to achieve higher performance.
These specialized algorithms target the unique challenges posed by different application domains, such as continuous control \citep{lillicrap2015ddpg}, discrete actions \citep{mnih2015dqn,schrittwieser2019muzero}, sparse rewards \citep{jaderberg2016unreal}, image inputs \citep{anand2019ataristaterep}, spatial environments \citep{driess2022nerfrl}, and board games \citep{silver2017alphago}.
However, applying reinforcement learning algorithms to sufficiently new tasks---such as moving from video games to robotics tasks---requires substantial effort, expertise, and computational resources for tweaking the hyperparameters of the algorithm \citep{andrychowicz2020whatmatters}.
This brittleness poses a bottleneck in applying reinforcement learning to new problems and also limits the applicability of reinforcement learning to computationally expensive models or tasks where tuning is prohibitive.
Creating a general algorithm that learns to master new domains without having to be reconfigured has been a central challenge in artificial intelligence and would open up reinforcement learning to a wide range of practical applications.

We present Dreamer, a general algorithm that outperforms specialized expert algorithms across a wide range of domains while using fixed hyperparameters, making reinforcement learning readily applicable to new problems.
The algorithm is based on the idea of learning a world model that equips the agent with rich perception and the ability to imagine the future\citep{sutton1991dyna,finn2017visualforesight,ha2018worldmodels}.
The world model predicts the outcomes of potential actions, a critic neural network judges the value of each outcome, and an actor neural network chooses actions to reach the best outcomes.
Although intuitively appealing, robustly learning and leveraging world models to achieve strong task performance has been an open problem\citep{kaiser2019simple}.
Dreamer overcomes this challenge through a range of robustness techniques based on normalization, balancing, and transformations.
We observe robust learning not only across over 150 tasks from the domains summarized in \cref{fig:tasks}, but also across model sizes and training budgets, offering a predictable way to increase performance.
Notably, larger model sizes not only achieve higher scores but also require less interaction to solve a task.

To push the boundaries of reinforcement learning, we consider the popular video game Minecraft that has become a focal point of research in recent years \citep{guss2019minerl,kanitscheider2021minecraftcurriculum,baker2022vpt}, with international competitions held for developing algorithms that autonomously learn to collect diamonds in Minecraft\footnote{
The MineRL Diamond Competitions were held in 2019, 2020, and 2021 and provided a dataset of human expert trajectories: \url{https://minerl.io/diamond}. Competitions in the following years focused on a wide range of tasks.}.
Solving this problem without human data has been widely recognized as a substantial challenge for artificial intelligence because of the sparse rewards, exploration difficulty, long time horizons, and the procedural diversity of this open world game \citep{guss2019minerl}.
Due to these obstacles, previous approaches resorted to using human expert data and domain-specific curricula \citep{kanitscheider2021minecraftcurriculum,baker2022vpt}.
Applied out of the box, Dreamer is the first algorithm to collect diamonds in Minecraft from scratch.

\begin{figure}[t!]
\centering
\vspace*{-1ex}
\begin{subfigure}{.5\textwidth}
\includegraphics[width=\linewidth]{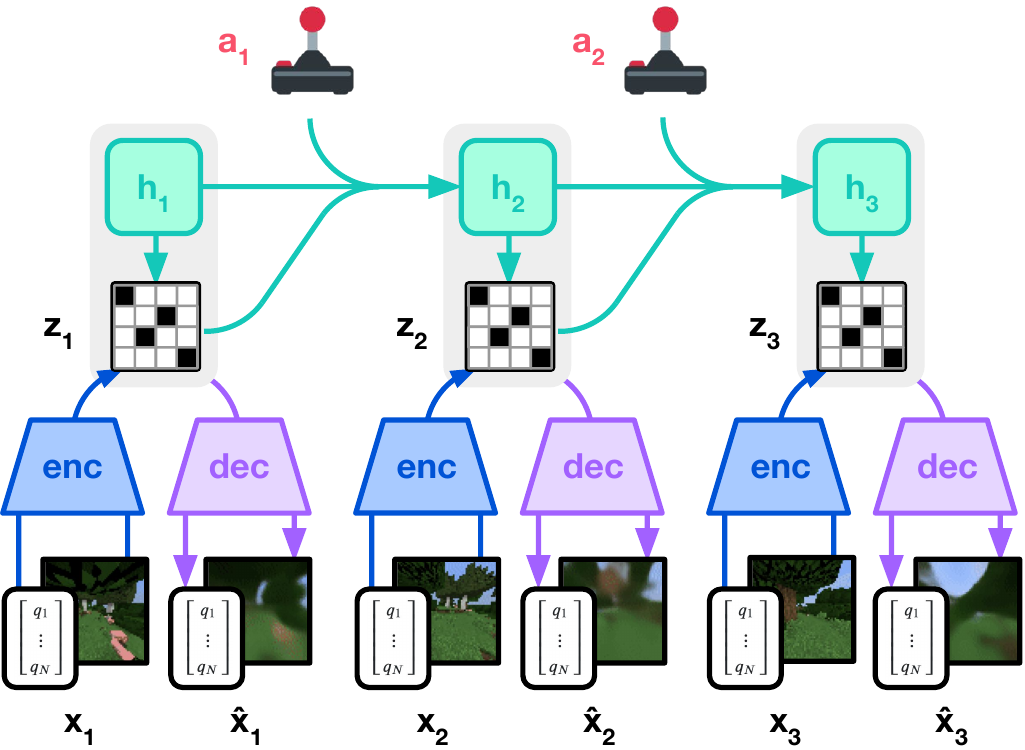}
\caption{World Model Learning}
\end{subfigure}\hfill%
\begin{subfigure}{.43\textwidth}
\includegraphics[width=\linewidth]{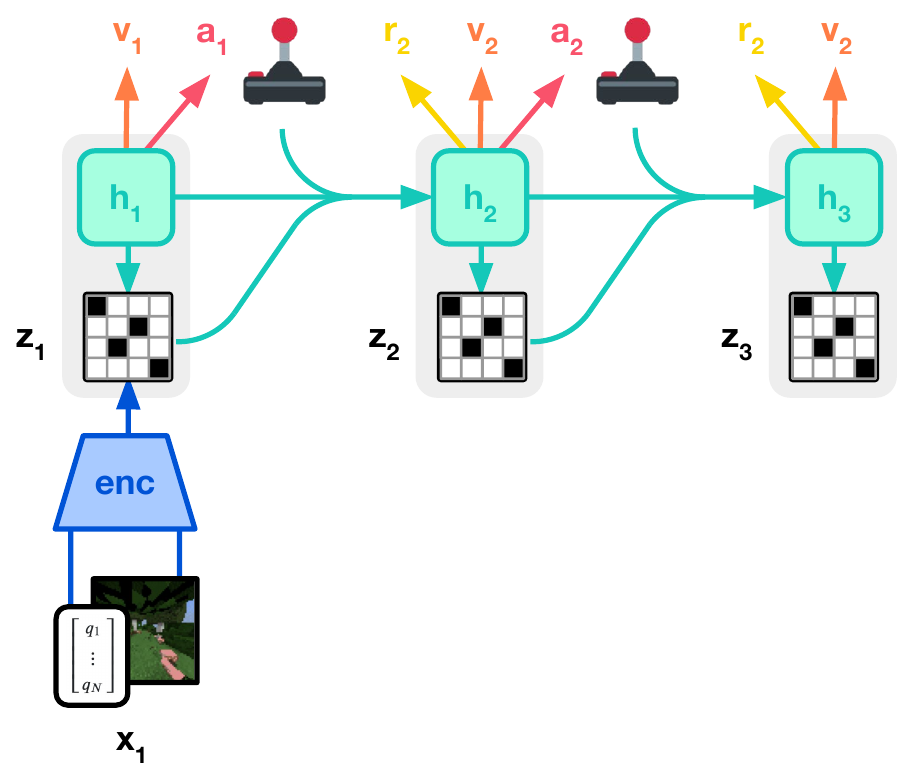}
\caption{Actor Critic Learning}
\end{subfigure}
\caption{Training process of Dreamer. The world model encodes sensory inputs into discrete representations $z_t$ that are predicted by a sequence model with recurrent state $h_t$ given actions $a_t$. The inputs are reconstructed to shape the representations. The actor and critic predict actions $a_t$ and values $v_t$ and learn from trajectories of abstract representations predicted by the world model.}
\label{fig:model}
\end{figure}

\section*{Learning algorithm}
\label{sec:algorithm}

We present the third generation of the Dreamer algorithm\citep{hafner2019dreamer,hafner2020dreamerv2}.
The algorithm consists of three neural networks: the world model predicts the outcomes of potential actions, the critic judges the value of each outcome, and the actor chooses actions to reach the most valuable outcomes.
The components are trained concurrently from replayed experience while the agent interacts with the environment.
To succeed across domains, all three components need to accommodate different signal magnitudes and robustly balance terms in their objectives.
This is challenging as we are not only targeting similar tasks within the same domain but aim to learn across diverse domains with fixed hyperparameters.
This section introduces the world model, critic, and actor along with their robust loss functions, as well as tools for robustly predicting quantities of unknown orders of magnitude.

\subsection*{World model learning}

The world model learns compact representations of sensory inputs through autoencoding \citep{kingma2013vae} and enables planning by predicting future representations and rewards for potential actions.
We implement the world model as a Recurrent State-Space Model (RSSM) \citep{hafner2018planet}, shown in \cref{fig:model}.
First, an encoder maps sensory inputs $x_t$ to stochastic representations $z_t$.
Then, a sequence model with recurrent state $h_t$ predicts the sequence of these representations given past actions $a_{t-1}$.
The concatenation of $h_t$ and $z_t$ forms the model state from which we predict rewards $r_t$ and episode continuation flags $c_t\in\{0,1\}$ and reconstruct the inputs to ensure informative representations:

\eq{
\begin{alignedat}{4}
\raisebox{1.95ex}{\llap{\blap{\ensuremath{
\text{RSSM} \hspace{1ex} \begin{cases} \hphantom{A} \\ \hphantom{A} \\ \hphantom{A} \end{cases} \hspace*{-2.4ex}
}}}}
& \text{Sequence model:}        \padspace && h_t            &\ =    &\ f_\phi(h_{t-1},z_{t-1},a_{t-1}) \\
& \text{Encoder:}   \padspace && z_t            &\ \sim &\ \qp(z_t | h_t,x_t) \\
& \text{Dynamics predictor:}   \padspace && \hat{z}_t      &\ \sim &\ \pp(\hat{z}_t | h_t) \\
& \text{Reward predictor:}       \padspace && \hat{r}_t      &\ \sim &\ \pp(\hat{r}_t | h_t,z_t) \\
& \text{Continue predictor:}     \padspace && \hat{c}_t      &\ \sim &\ \pp(\hat{c}_t | h_t,z_t) \\
& \text{Decoder:}        \padspace && \hat{x}_t      &\ \sim &\ \pp(\hat{x}_t | h_t,z_t)
\end{alignedat}}

\begin{figure}[t]
\centering
\vspace*{-3ex}
\includegraphics[width=1\linewidth,trim={0 .5cm 0 0},clip]{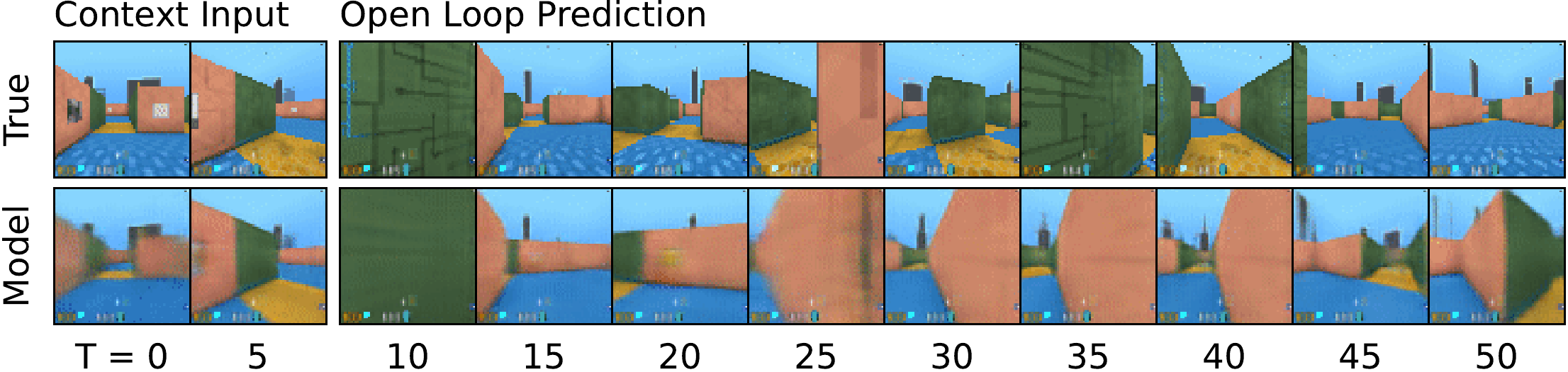} \\[.5ex]
\includegraphics[width=1\linewidth,trim={0 0 0 .5cm},clip]{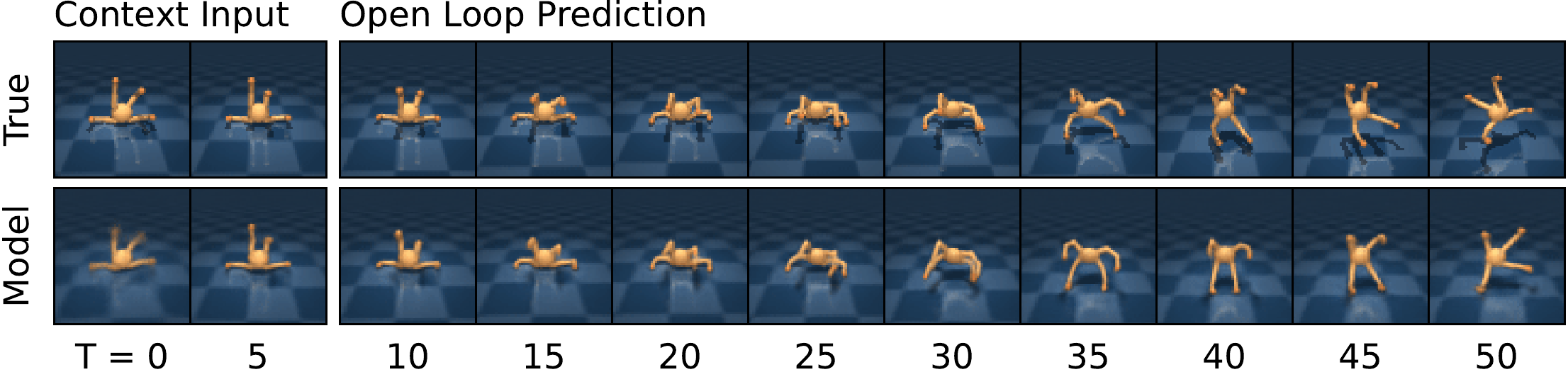}
\caption{Multi-step video predictions of a DMLab maze (top) and a quadrupedal robot (bottom). Given 5 context images and the full action sequence, the model predicts 45 frames into the future without access to intermediate images. The world model learns an understanding of the underlying structure of each environment.}
\label{fig:openl}
\end{figure}

\Cref{fig:openl} visualizes long-term video predictions of the world world.
The encoder and decoder use convolutional neural networks (CNN) for image inputs and multi-layer perceptrons (MLPs) for vector inputs.
The dynamics, reward, and continue predictors are also MLPs.
The representations are sampled from a vector of softmax distributions and we take straight-through gradients through the sampling step \citep{bengio2013straight,hafner2020dreamerv2}.
Given a sequence batch of inputs $x_{1:T}$, actions $a_{1:T}$, rewards $r_{1:T}$, and continuation flags $c_{1:T}$, the world model parameters $\phi$ are optimized end-to-end to minimize the prediction loss $\mathcal{L}_{\mathrm{pred}}$, the dynamics loss $\mathcal{L}_{\mathrm{dyn}}$, and the representation loss $\mathcal{L}_{\mathrm{rep}}$ with corresponding loss weights $\beta_{\mathrm{pred}}=1$, $\beta_{\mathrm{dyn}}=1$, and $\beta_{\mathrm{rep}}=0.1$:

\eq{
\mathcal{L}(\phi)\doteq
\E_{q_\phi}<\Big>[\textstyle\sum_{t=1}^T(
    \beta_{\mathrm{pred}}\mathcal{L}_{\mathrm{pred}}(\phi)
   +\beta_{\mathrm{dyn}}\mathcal{L}_{\mathrm{dyn}}(\phi)
   +\beta_{\mathrm{rep}}\mathcal{L}_{\mathrm{rep}}(\phi)
)].
\label{eq:wm}
}

The prediction loss trains the decoder and reward predictor via the symlog squared loss described later, and the continue predictor via logistic regression.
The dynamics loss trains the sequence model to predict the next representation by minimizing the KL divergence between the predictor $\pp(z_t|h_t)$ and the next stochastic representation $\qp(z_t|h_t,x_t)$.
The representation loss, in turn, trains the representations to become more predictable allowing us to use a factorized dynamics predictor for fast sampling during imagination training.
The two losses differ in the stop-gradient operator $\operatorname{sg}(\cdot)$ and their loss scale.
To avoid a degenerate solution where the dynamics are trivial to predict but fail to contain enough information about the inputs, we employ free bits\citep{kingma2016freebits} by clipping the dynamics and representation losses below the value of 1 nat $\approx$ 1.44 bits.
This disables them while they are already minimized well to focus learning on the prediction loss:

\eq{
\mathcal{L_{\mathrm{pred}}}(\phi) & \doteq
-\lnpp(x_t|z_t,h_t)
-\lnpp(r_t|z_t,h_t)
-\lnpp(c_t|z_t,h_t) \\
\mathcal{L_{\mathrm{dyn}}}(\phi) & \doteq
\max\bigl(1,\KL[\sg(\qp(z_t|h_t,x_t)) || \hspace{3.2ex}\pp(z_t|h_t)\hphantom{)}]\bigr) \\
\mathcal{L_{\mathrm{rep}}}(\phi) & \doteq
\max\bigl(1,\KL[\hspace{3.2ex}\qp(z_t|h_t,x_t)\hphantom{)} || \sg(\pp(z_t|h_t))]\bigr)
}

Previous world models require scaling the representation loss differently based on the visual complexity of the environment \citep{hafner2019dreamer}.
Complex 3D environments contain details unnecessary for control and thus prompt a stronger regularizer to simplify the representations and make them more predictable.
In games with static backgrounds and where individual pixels may matter for the task, a weak regularizer is required to extract fine details.
We find that combining free bits with a small representation loss resolves this dilemma, allowing for fixed hyperparameters across domains.
Moreover, we transform vector observations using the symlog function described later, to prevent large inputs and large reconstruction gradients, further stabilizing the trade-off with the representation loss.

We occasionally observed spikes the in KL losses in earlier experiments, consistent with reports for deep variational autoencoders \citep{child2020vdvae}.
To prevent this, we parameterize the categorical distributions of the encoder and dynamics predictor as mixtures of 1\% uniform and 99\% neural network output, making it impossible for them to become deterministic and thus ensuring well-behaved KL losses.
Further model details and hyperparameters are included in the supplementary material.

\subsection*{Critic learning}

The actor and critic neural networks learn behaviors purely from abstract trajectories of representations predicted by the world model \citep{sutton1991dyna}.
For environment interaction, we select actions by sampling from the actor network without lookahead planning.
The actor and critic operate on model states $s_t \doteq \{h_t,z_t\}$ and thus benefit from the Markovian representations learned by the recurrent world model.
The actor aims to maximize the return $R_t \doteq \textstyle\sum_{\tau=0}^\infty \gamma^\tau r_{t+\tau}$ with a discount factor $\gamma=0.997$ for each model state.
To consider rewards beyond the prediction horizon $T=16$, the critic learns to approximate the distribution of returns \citep{bellemare2017c51} for each state under the current actor behavior:

\eq{
&\text{Actor:}\quad
&& a_t \sim \p<\pi_\theta>(a_t|s_t) \qquad
&\text{Critic:}\quad
&& \p<v_\psi>(R_t | s_t)
\label{eq:ac}
}

Starting from representations of replayed inputs, the world model and actor generate a trajectory of
imagined model states $s_{1:T}$, actions $a_{1:T}$, rewards $r_{1:T}$, and continuation flags $c_{1:T}$.
Because the critic predicts a distribution, we read out its predicted values $v_t \doteq \operatorname{E}[\p<v_\psi>(\,\cdot|s_t)]$ as the expectation of the distribution.
To estimate returns that consider rewards beyond the prediction horizon, we compute bootstrapped $\lambda$-returns \citep{sutton2018rlbook} that integrate the predicted rewards and the values.
The critic learns to predict the distribution of the return estimates $R^\lambda_t$ using the maximum likelihood loss:

\eq{
\mathcal{L}(\psi) \doteq -\textstyle\sum_{t=1}^T \p<\ln p_\psi>(R^\lambda_t | s_t)
\qquad
R^\lambda_t \doteq r_t + \gamma c_t \Big(
  (1 - \lambda) v_t +
  \lambda R^\lambda_{t+1}
\Big)
\qquad
R^\lambda_T \doteq v_T
\label{eq:lambda}
}

While a simple choice would be to parameterize the critic as a Normal distribution, the return distribution can have multiple modes and vary by orders of magnitude across environments. To stabilize and accelerate learning under these conditions, we parameterize the critic as categorical distribution with exponentially spaced bins, decoupling the scale of gradients from the prediction targets as described later.
To improve value prediction in environments where rewards are challenging to predict, we apply the critic loss both to imagined trajectories with loss scale $\beta_{\mathrm{val}}=1$ and to trajectories sampled from the replay buffer with loss scale $\beta_{\mathrm{repval}}=0.3$. The critic replay loss uses the imagination returns $R^\lambda_t$ at the start states of the imagination rollouts as on-policy value annotations for the replay trajectory to then compute $\lambda$-returns over the replay rewards.

Because the critic regresses targets that depend on its own predictions, we stabilize learning by regularizing the critic towards predicting the outputs of an exponentially moving average of its own parameters.
This is similar to target networks used previously in reinforcement learning \citep{mnih2015dqn} but allows us to compute returns using the current critic network.
We further noticed that the randomly initialized reward predictor and critic networks at the start of training can result in large predicted rewards that can delay the onset of learning.
We thus initialize the output weight matrix of the reward predictor and critic to zeros, which alleviates the problem and accelerates early learning.

\subsection*{Actor learning}

The actor learns to choose actions that maximize return while exploring through an entropy regularizer \citep{williams1991maxentreinforce}.
However, the correct scale for this regularizer depends both on the scale and frequency of rewards in the environment.
Ideally, we would like the agent to explore more if rewards are sparse and exploit more if rewards are dense or nearby. At the same time, the exploration amount should not be influenced by arbitrary scaling of rewards in the environment.
This requires normalizing the return scale while preserving information about reward frequency.

To use a fixed entropy scale of $\eta=3\times10^{-4}$ across domains, we normalize returns to be approximately contained in the interval $[0, 1]$. In practice, substracting an offset from the returns does not change the actor gradient and thus dividing by the range $S$ is sufficient. Moreover, to avoid amplifying noise from function approximation under sparse rewards, we only scale down large return magnitudes but leave small returns below the threshold of $L=1$ untouched. We use the Reinforce estimator \citep{williams1992reinforce} for both discrete and continuous actions, resulting in the surrogate loss function:

\eq{
\mathcal{L}(\theta)\doteq
-\textstyle\sum_{t=1}^T
\sg\!\Big(\big(R^\lambda_t-v_\psi(s_t)\big)/\max(1,\,S)\Big)
\log\p<\pi_\theta>(a_t|s_t)
+\,\eta\H[\p<\pi_\theta>(a_t|s_t)]
}

The return distribution can be multi-modal and include outliers, especially for randomized environments where some episodes have higher achievable returns than others. Normalizing by the smallest and largest observed returns would then scale returns down too much and may cause suboptimal convergence. To be robust to these outliers, we compute the range from the 5\textsuperscript{th} to the 95\textsuperscript{th} return percentile over the return batch and smooth out the estimate using an exponential moving average:

\eq{
S \doteq \EMA\!\big(\Per(R^\lambda_t, 95) - \Per(R^\lambda_t, 5), 0.99\big)
}

Previous work typically normalizes advantages\citep{schulman2017ppo} rather than returns, which puts a fixed amount of emphasis on maximizing returns over entropy regardless of whether rewards are within reach.
Scaling up advantages when rewards are sparse can amplify noise that outweighs the entropy regularizer and stagnates exploration.
Normalizing rewards or returns by standard deviation can fail under sparse rewards where their standard deviation is near zero, drastically amplifying rewards regardless of their size.
Constrained optimization targets a fixed entropy on average across states \citep{haarnoja2018sac,abdolmaleki2018mpo} regardless of achievable returns, which is robust but explores slowly under sparse rewards and converges lower under dense rewards.
We did not find stable hyperparameters across domains for these approaches.
Return normalization with a denominator limit overcomes these challenges, exploring rapidly under sparse rewards and converging to high performance across diverse domains.

\subsection*{Robust predictions}
\label{sec:symlog}

Reconstructing inputs and predicting rewards and returns can be challenging because the scale of these quantities can vary across domains.
Predicting large targets using a squared loss can lead to divergence whereas absolute and Huber losses \citep{mnih2015dqn} stagnate learning.
On the other hand, normalizing targets based on running statistics \citep{schulman2017ppo} introduces non-stationarity into the optimization.
We suggest the symlog squared error as a simple solution to this dilemma.
For this, a neural network $f(x,\theta)$ with inputs $x$ and parameters $\theta$ learns to predict a transformed version of its targets $y$.
To read out predictions $\hat{y}$ of the network, we apply the inverse transformation:

\eq{
\mathcal{L(\theta)} \doteq \textstyle\frac{1}{2}\big(f(x,\theta)-\symlog(y)\big)^2 \qquad
\hat{y} \doteq \symexp\!\big(f(x,\theta)\big)
\label{eq:logpred}}

Using the logarithm as transformation would not allow us to predict targets that take on negative values.
Therefore, we choose a function from the bi-symmetric logarithmic family \citep{webber2012symlog} that we name symlog as the transformation with the symexp function as its inverse:

\eq{
\symlog(x) \doteq \sign(x)\ln\!\big(|x|+1\big) \qquad
\symexp(x) \doteq \sign(x)\big(\!\exp(|x|)-1\big)
\label{eq:symlog}}

The symlog function compresses the magnitudes of both large positive and negative values.
Unlike the logarithm, it is symmetric around the origin while preserving the input sign.
This allows the optimization process to quickly move the network predictions to large values when needed.
The symlog function approximates the identity around the origin so that it does not affect learning of targets that are already small enough.

For potentially stochastic targets, such as rewards or returns, we introduce the symexp twohot loss. Here, the network outputs the logits for a softmax distribution over exponentially spaced bins $b_i \in B$.
Predictions are read out as the weighted average of the bin positions weighted by their predicted probabilities. Importantly, the network can output any continuous value in the interval because the weighted average can fall between the buckets:

\eq{
\hat{y} \doteq \operatorname{softmax}(f(x))^T B \qquad
B \doteq \symexp(\begin{bmatrix} -20 & ... & +20 \end{bmatrix})
}

The network is trained on twohot encoded targets \citep{schrittwieser2019muzero,bellemare2017c51}, a generalization of onehot encoding to continuous values. The twohot encoding of a scalar is a vector with $|B|$ entries that are all $0$ except at the indices $k$ and $k+1$ of the two bins closest to the encoded scalar. The two entries sum up to $1$, with linearly higher weight given to the bin that is closer to the encoded continuous number.
The network is then trained to minimize the categorical cross entropy loss for classification with soft targets. Note that the loss only depends on the probabilities assigned to the bins but not on the continuous values associated with the bin locations, decoupling the size of the gradients from the size of the targets:

\eq{
\mathcal{L}(\theta) \doteq
-\twohot(y)^T \operatorname{log\,softmax}(f(x,\theta))
\label{eq:twohotloss}
}

Applying these principles, Dreamer transforms vector observations using the symlog functions, both for the encoder inputs and the decoder targets and employs the synexp twohot loss for the reward predictor and critic.
We find that these techniques enable robust and fast learning across many diverse domains.
For critic learning, an alternative asymmetric transformation has previously been proposed \citep{kapturowski2018r2d2}, which we found less effective on average across domains.
Unlike alternatives, symlog transformations avoid truncating large targets \citep{mnih2015dqn}, introducing non-stationary from normalization \citep{schulman2017ppo}, or adjusting network weights when new extreme values are detected \citep{hessel2019popart}.

\begin{figure}[t]
\centering
\includegraphics[width=\linewidth]{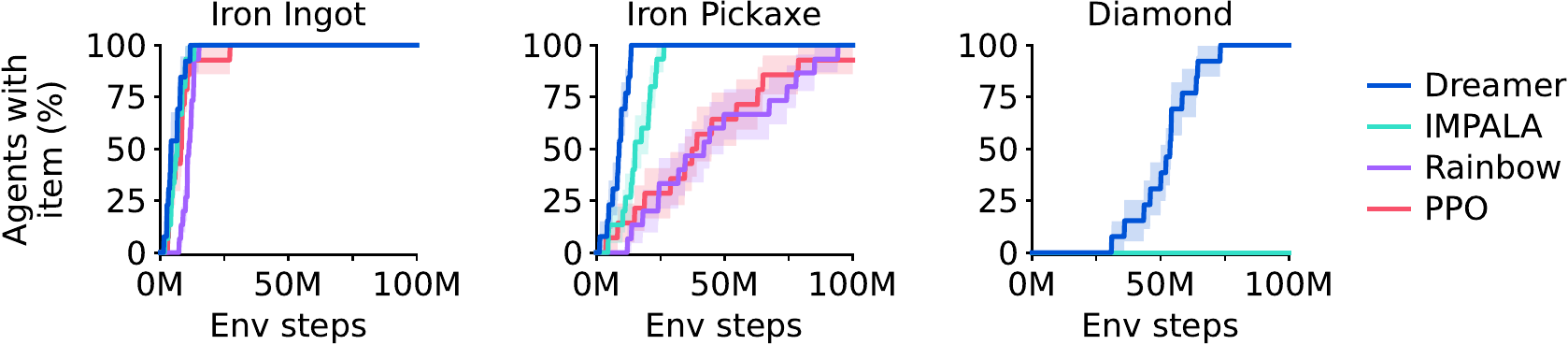}%
\caption{Fraction of trained agents that discover each of the three latest items in the Minecraft Diamond task. Although previous algorithms progress up to the iron pickaxe, Dreamer is the only compared algorithm that manages to discover a diamond, and does so reliably.}
\label{fig:minecraft}
\end{figure}

\begin{figure}[t]
\centering
\begin{subfigure}{\textwidth}
\includegraphics[width=\linewidth]{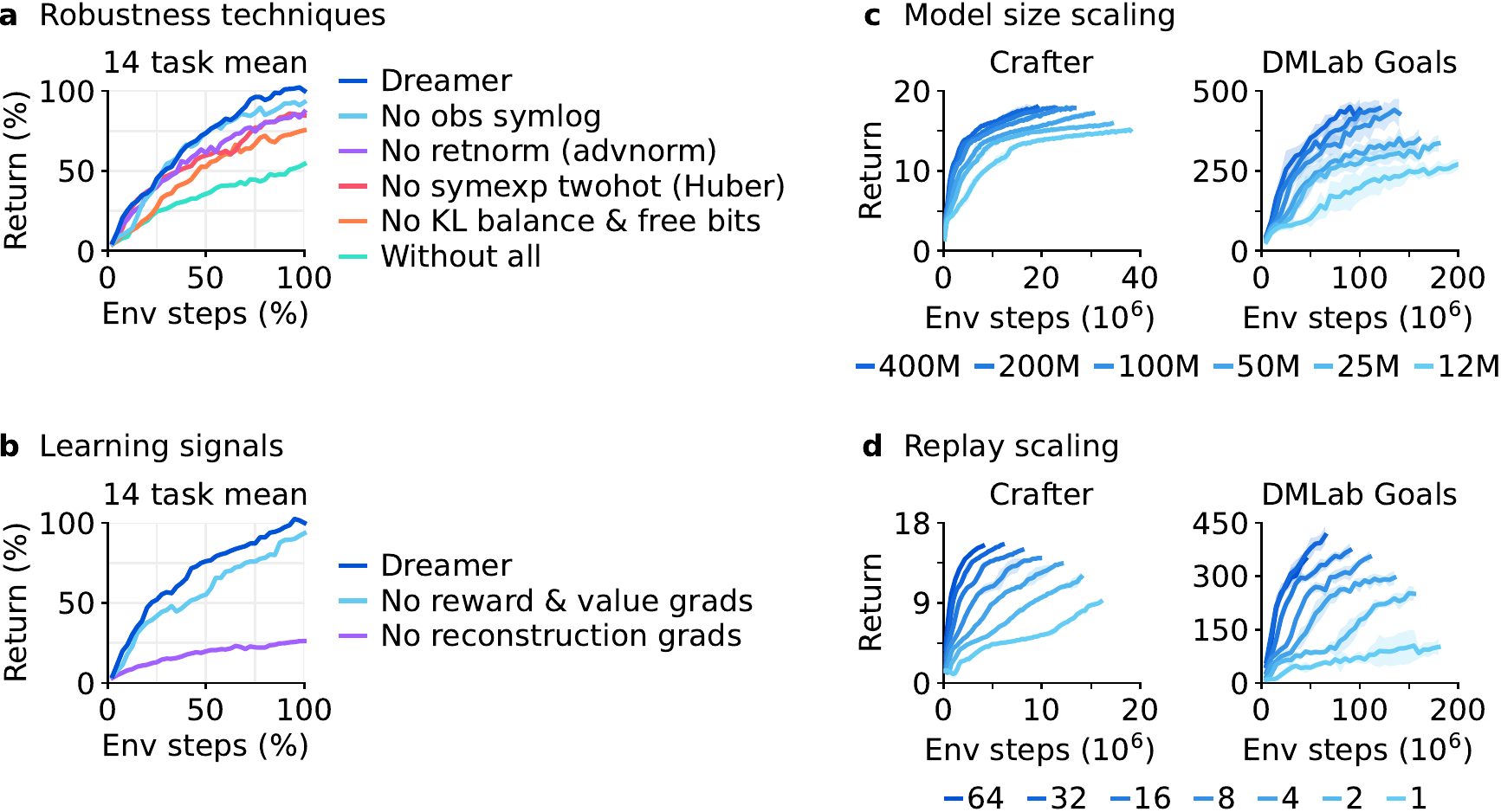}
\end{subfigure}%
\caption{Ablations and robust scaling of Dreamer.
\textbf{a}, All individual robustness techniques contribute to the performance of Dreamer on average, although each individual technique may only affect some tasks. Training curves of individual tasks are included in the supplementary material.
\textbf{b}, The performance of Dreamer predominantly rests on the unsupervised reconstruction loss of its world model, unlike most prior algorithms that rely predominantly on reward and value prediction gradients\citep{mnih2015dqn,schulman2017ppo,schrittwieser2019muzero}.
\textbf{c}, The performance of Dreamer increases monotonically with larger model sizes, ranging from 12M to 400M parameters. Notably, larger models not only increase task performance but also require less environment interaction.
\textbf{d}, Higher replay ratios predictably increase the performance of Dreamer. Together with model size, this allows practitioners to improve task performance and data-efficiency by employing more computational resources.
}
\label{fig:ablations}
\end{figure}

\section*{Results}
\label{sec:experiments}

We evaluate the generality of Dreamer across 8 domains---with over 150 tasks---under fixed hyperparameters.
We designed the experiments to compare Dreamer to the best methods in the literature, which are often specifically designed and tuned for the benchmark at hand.
We further compare to a high-quality implementation of PPO\citep{schulman2017ppo}, a standard reinforcement learning algorithm that is known for its robustness. We run PPO with fixed hyperparameters chosen to maximize performance across domains and that reproduce strong published results of PPO on ProcGen\citep{cobbe2021ppg}.
To push the boundaries of reinforcement learning, we apply Dreamer to the challenging video game Minecraft, comparing it to strong previous algorithms.
Finally, we analyze the importance of individual components of Dreamer and its robustness to different model sizes and computational budgets.
All Dreamer agents are trained on a single Nvidia A100 GPU each, making it reproducible for many research labs.
A public implementation of Dreamer that reproduces all results is available on the project website.

\paragraph{Benchmarks}
We perform an extensive empirical study across 8 domains that include continuous and discrete actions, visual and low-dimensional inputs, dense and sparse rewards, different reward scales, 2D and 3D worlds, and procedural generation.
\Cref{fig:bars} summarizes the benchmark results, showing that Dreamer outperforms a wide range of previous expert algorithms across diverse domains. Crucially, Dreamer substantially outperforms PPO across all domains.

\begin{itemize}
\itempar{Atari} This established benchmark contains 57 Atari 2600 games with a budget of 200M frames, posing a diverse range of challenges\citep{bellemare2013ale}.
We use the sticky action simulator setting \citep{machado2018revisiting}.
Dreamer outperforms the powerful MuZero algorithm\citep{schrittwieser2019muzero} while using only a fraction of the computational resources.
Dreamer also outperforms the widely-used expert algorithms Rainbow \citep{hessel2018rainbow} and IQN \citep{dabney2018iqn}.
\itempar{ProcGen} This benchmark of 16 games features randomized levels and visual distractions to test the robustness and generalization of agents\citep{cobbe2020procgen}. Within the budget of 50M frames, Dreamer matches the tuned expert algorithm PPG \citep{cobbe2021ppg} and outperforms Rainbow\citep{cobbe2020procgen,hessel2018rainbow}. Our PPO agent with fixed hyperparameters matches the published score of the highly tuned official PPO implementation\citep{cobbe2021ppg}.
\itempar{DMLab} This suite of 30 tasks features 3D environments that test spatial and temporal reasoning \citep{beattie2016dmlab}. In 100M frames, Dreamer exceeds the performance of the scalable IMPALA and R2D2+ agents \citep{kapturowski2018r2d2} at 1B environment steps, amounting to a data-efficiency gain of over 1000\%.
We note that these baselines were not designed for data-efficiency but serve as a valuable comparison point for the performance previously achievable at scale.
\itempar{Atari100k} This data-efficiency benchmark comntains 26 Atari games and a budget of only 400K frames, amounting to 2 hours of game time \citep{kaiser2019simple}. EfficientZero \citep{ye2021effzero} holds the state-of-the-art by combining online tree search, prioritized replay, and hyperparameter scheduling, but also resets levels early to increase data diversity, making a comparison difficult.
Without this complexity, Dreamer outperforms the best remaining methods, including the transformer-based IRIS and TWM agents, the model-free SPR, and SimPLe \citep{micheli2022iris}.
\itempar{Proprio Control} This benchmark contains 18 control tasks with continuous actions, proprioceptive vector inputs, and a budget of 500K environment steps \citep{tassa2018dmcontrol}. The tasks range from classical control over locomotion to robot manipulation tasks, featuring dense and sparse rewards. Dreamer sets a new state-of-the-art on this benchmark,
outperforming D4PG, DMPO, and MPO \citep{abdolmaleki2018mpo}.
\itempar{Visual Control} This benchmark consists of 20 continuous control tasks where the agent receives only high-dimensional images as input and has a budget of 1M environment steps \citep{tassa2018dmcontrol}.
Dreamer establishes a new state-of-the-art on this benchmark, outperforming DrQ-v2 and CURL \citep{yarats2021drqv2}, which are specialized to visual environments and leverage data augmentation.
\pagebreak
\itempar{BSuite} This benchmark includes 23 environments with a total of 468 configurations that are specifically designed to test credit assignment, robustness to reward scale and stochasticity, memory, generalization, and exploration \citep{osband2019bsuite}. Dreamer establishes a new state-of-the-art on this benchmark, outperforming Boot DQN and other methods \citep{dizon2023bsuiteagents}.
Dreamer improves over previous algorithms especially in the scale robustness category.
\end{itemize}

\paragraph{Minecraft}
Collecting diamonds in the popular game Minecraft has been a long-standing challenge in artificial intelligence\citep{guss2019minerl,kanitscheider2021minecraftcurriculum,baker2022vpt}.
Every episode in this game is set in a unique randomly generated and infinite 3D world. Episodes last until the player dies or up to 36000 steps equaling 30 minutes, during which the player needs to discover a sequence of 12 items from sparse rewards by foraging for resources and crafting tools. It takes about 20 minutes for experienced human players to obtain diamonds\citep{baker2022vpt}.
We follow the block breaking setting of prior work \citep{kanitscheider2021minecraftcurriculum} because the provided action space would make it challenging for stochastic policies to keep a key pressed for a prolonged time.

Because of the training time in this complex domain, extensive tuning would be difficult for Minecraft.
Instead, we apply Dreamer out of the box with its default hyperparameters.
As shown in \cref{fig:minecraft,fig:bars}, Dreamer is the first algorithm to collect diamonds in Minecraft from scratch without using human data as was required by VPT \citep{baker2022vpt} or adaptive curricula\citep{kanitscheider2021minecraftcurriculum}.
All the Dreamer agents we trained on Minecraft discover diamonds in 100M environment steps.
While several strong baselines progress to advanced items such as the iron pickaxe, none of them discovers a diamond.

\paragraph{Ablations}
In \cref{fig:ablations}, we ablate the robustness techniques and learning signals on a diverse set of 14 tasks to understand their importance.
The training curves of individual tasks are included in the supplementary material.
We observe that all robustness techniques contribute to performance, most notably the KL objective of the world model, followed by return normalization and symexp twohot regression for reward and value prediction. In general, we find that each individual technique is critical on a subset of tasks but may not affect performance on other tasks.

To investigate the effect of the world model, we ablate the learning signals of Dreamer by stopping either the task-specific reward and value prediction gradients or the task-agnostic reconstruction gradients from shaping its representations. Unlike previous reinforcement learning algorithms that often rely only on task-specific learning signals\citep{mnih2015dqn,schrittwieser2019muzero}, Dreamer rests predominantly on the unsupervised objective of its world model. This finding could allow for future algorithm variants that leverage pretraining on unsupervised data.

\paragraph{Scaling properties}
To investigate whether Dreamer can scale robustly, we train 6 model sizes ranging from 12M to 400M parameters, as well as different replay ratios on Crafter \citep{hafner2021crafter} and a DMLab task\citep{beattie2016dmlab}. The replay ratio affects the number of gradient updates performed by the agent.
\Cref{fig:ablations} shows robust learning with fixed hyperparameters across the compared model sizes and replay ratios.
Moreover, increasing the model size directly translates to both higher task performance and a lower data requirement.
Increasing the number of gradient steps further reduces the interactions needed to learn successful behaviors.
The results show that Dreamer learns robustly across model sizes and replay ratios and that its performance and provides a predictable way for increasing performance given computational resources.  

\pagebreak
\section*{Previous work}
\label{sec:related}

Developing general-purpose algorithms has long been a goal of reinforcement learning research.
PPO \citep{schulman2017ppo} is one of the most widely used algorithms and is relatively robust but requires large amounts of experience and often yields lower performance than specialized alternatives.
SAC \citep{haarnoja2018sac} is a popular choice for continuous control and leverages experience replay for data-efficiency, but in practice requires tuning, especially for its entropy scale, and struggles under high-dimensional inputs \citep{yarats2019sacae}.
MuZero \citep{schrittwieser2019muzero} plans using a value prediction model and has been applied to board games and Atari, but the authors did not release an implementation and the algorithm contains several complex components, making it challenging to reproduce.
Gato \citep{reed2022gato} fits one large model to expert demonstrations of multiple tasks, but is only applicable when expert data is available.
In comparison, Dreamer masters a diverse range of environments with fixed hyperparameters, does not require expert data, and its implementation is open source.

Minecraft has been a focus of recent research. With MALMO \citep{johnson2016malmo}, Microsoft released a free version of the successful game for research purposes.
MineRL \citep{guss2019minerl} offers several competition environments, which we rely on as the basis for our experiments.
The MineRL competition supports agents in exploring and learning meaningful skills through a diverse human dataset \citep{guss2019minerl}.
Voyager obtains items at a similar depth in the technology tree as Dreamer using API calls to a language model but operates on top of the MineFlayer bot scripting layer that was specifically engineered to the game and exposes high-level actions\citep{wang2023voyager}.
VPT \citep{baker2022vpt} trained an agent to play Minecraft through behavioral cloning based on expert data of keyboard and mouse actions collected by contractors and finetuning using reinforcement learning to obtain diamonds using 720 GPUs for 9 days.
In comparison, Dreamer uses the MineRL competition action space to autonomously learn to collect diamonds from sparse rewards using 1 GPU for 9 days, without human data.

\section*{Conclusion}
\label{sec:discussion}

We present the third generation of the Dreamer algorithm, a general reinforcement learning algorithm that masters a wide range of domains with fixed hyperparameters.
Dreamer excels not only across over 150 tasks but also learns robustly across varying data and compute budgets, moving reinforcement learning toward a wide range of practical applications.
Applied out of the box, Dreamer is the first algorithm to collect diamonds in Minecraft from scratch, achieving a significant milestone in the field of artificial intelligence.
As a high-performing algorithm that is based on a learned world model, Dreamer paves the way for future research directions, including teaching agents world knowledge from internet videos and learning a single world model across domains to allow artificial agents to build up increasingly general knowledge and competency.

\paragraph{Acknowledgements}
We thank Mohammad Norouzi, Jessy Lin, Abbas Abdolmaleki, John Schulman, Adam Kosiorek, and Oleh Rybkin for insightful discussions.
We thank Bobak Shahriari, Denis Yarats, Karl Cobbe,
and Hubert Soyer for sharing training curves of baseline algorithms. We thank Daniel Furrer, Andrew Chen, and Dakshesh Garambha for help with Google Cloud infrastructure.

\clearpage
\begin{hyphenrules}{nohyphenation}
\setlength{\bibsep}{.5ex plus .8ex}
\bibliography{references}
\end{hyphenrules}

\clearpage
\appendix
\clearpage

\section*{Methods}

\subsection*{Baselines}

We employ the Proximal Policy Optimization (PPO) algorithm \citep{schulman2017ppo}, which has become a standard choice in the field, to compare Dreamer under fixed hyperparameters across all benchmarks. There are a large number of PPO implementations available publicly and they are known to substantially vary in task performance \citep{huang2022ppodetails}. To ensure a comparison that is representative of the highest performance PPO can achieve under fixed hyperparameters across domains, we choose the high-quality PPO implementation available in the Acme framework \citep{hoffman2020acme} and select its hyperparameters in \cref{tab:hparams_ppo} following recommendations \citep{huang2022ppodetails,andrychowicz2020whatmatters} and additionally tune its epoch batch size to be large enough for complex environments \cite{cobbe2020procgen}, its learning rate, and its entropy scale. We match the discount factor to Dreamer because it works well across domains and is a common choice in the literature \citep{kapturowski2018r2d2,schrittwieser2019muzero}. We choose the IMPALA network architecture that we have found performed better than alternatives \citep{cobbe2020procgen} and set the minibatch size to the largest possible for one A100 GPU. We verify the performance of our PPO implementation and hyperparameters on the ProcGen benchmark, where a highly tuned PPO implementation has been reported by the PPO authors \citep{cobbe2021ppg}. We find that our implementation matches or slightly outperforms this performance reference.

\begin{table}[h!]
\centering
\begin{mytabular}{
  colspec = {| L{12em} | C{5em} |},
  row{1} = {font=\bfseries},
}

\toprule
\textbf{Parameter} & \textbf{Value} \\
\midrule
Observation normalization & Yes \\
Reward normalization & Yes \\
Reward clipping (stddev.) & 10 \\
Epoch batch & $64 \times 256$ \\
Number of epochs & 3 \\
Minibatch size & 8 \\
Minibatch length & 256 \\
Policy trust region & 0.2 \\
Value trust region & No \\
Advantage normalization & Yes \\
Entropy penalty scale & 0.01 \\
Discount factor & 0.997 \\
GAE lambda & 0.95 \\
Learning rate & $3 \times 10^{-4}$ \\
Gradient clipping & $0.5$ \\
Adam epsilon & $10^{-5}$ \\
\bottomrule

\end{mytabular}
\caption{PPO hyperparameters used across all benchmarks.
}
\label{tab:hparams_ppo}
\end{table}

For Minecraft, we additionally tune and run the IMPALA and Rainbow algorithms because not successful end-to-end learning from scratch has been reported in the literature\citep{guss2019minerl}. We use the Acme implementations\citep{hoffman2020acme} of these algorithms, use the same IMPALA network we used for PPO, and tuned the learning rate and entropy regularizers.
For all other benchmarks, we compare to tuned expert algorithms reported in the literature as referenced in the results section.

\vspace*{-3ex}
\subsection*{Implementation}

\paragraph{Experience replay}
We implement Dreamer using a uniform replay buffer with an online queue\citep{schmitt2020laser}. Specifically, each minibatch is formed first from non-overlapping online trajectories and then filled up with uniformly sampled trajectories from the replay buffer. We store latent states into the replay buffer during data collection to initialize the world model on replayed trajectories, and write the fresh latent states of the training rollout back into the buffer.
While prioritized replay\citep{schaul2015prioritized} is used by some of the expert algorithms we compare to and we found it to also improve the performance of Dreamer, we opt for uniform replay in our experiments for ease of implementation.

We parameterize the amount of training via the replay ratio. This is the fraction of time steps trained on per time step collected from the environment, without action repeat. Dividing the replay ratio by the time steps in a minibatch and by action repeat yields the ratio of gradient steps to env steps. For example, a replay ratio of 32 on Atari with action repeat of 4 and batch shape $16 \times 64$ corresponds to 1 gradient step every 128 env steps, or 1.5M gradient steps over 200M env steps.

\paragraph{Optimizer}
We employ Adaptive Gradient Clipping (AGC) \citep{brock2021agc}, which clips per-tensor gradients if they exceed $30\%$ of the L2 norm of the weight matrix they correspond to, with its default $\eps=10^{-3}$. AGC decouples the clipping threshold from the loss scales, allowing to change loss functions or loss scales without adjusting the clipping threshold. We apply the clipped gradients using the LaProp optimizer\citep{ziyin2020laprop} with $\eps=10^{-20}$ and its default parameters $\beta_1=0.9$ and $\beta_2=0.99$. LaProp normalizes gradients by RMSProp and then smoothes them by momentum, instead of computing both momentum and normalizer on raw gradients as Adam does\citep{kingma2014adam}. This simple change allows for a smaller epsilon and avoids occasional instabilities that we observed under Adam.

\paragraph{Distributions}
The encoder, dynamics predictor, and actor distributions are mixtures of 99\% the predicted softmax output and 1\% of a uniform distribution \citep{gruslys2017reactor} to prevent zero probabilities and infinite log probabilities.
The rewards and critic neural networks output a softmax distribution over exponentially spaced bins $b \in B$ and are trained towards twohot encoded targets:

\eq{
\operatorname{twohot}(x)_i
\doteq \begin{cases}
|b_{k+1} - x| \,/\, |b_{k+1} - b_k| &\text{ if } i=k \\
|b_{k\hphantom{+1}} - x| \,/\, |b_{k+1} - b_k| &\text{ if } i=k+1 \\
0 &\text{ else} \\
\end{cases} \qquad
k \doteq \sum_{j=1}^{|B|} \operatorname{\delta}(b_j < x)
\label{eq:twohot}
}

The output weights of twohot distributions are initialized to zero to ensure that the agent does not hallucinate rewards and values at initialization. For computing the expected prediction of the softmax distribution under bins that span many orders of magnitude, the summation order matters and positive and negative bins should be summed up separately, from small to large bins, and then added. Refer to the source code for an implementation.

\paragraph{Networks}
Images are encoded using stride 2 convolutions to resolution $6 \times 6$ or $4 \times 4$ and then flattened and are decoded using transposed stride 2 convolutions, with sigmoid activation on the output. Vector inputs are symlog transformed and then encoded and decoded using 3-layer MLPs. The actor and critic neural networks are also 3-layer MLPs and the reward and continue predictors are 1-layer MLPs.
The sequence model is a GRU\citep{cho2014gru} with block-diagonal recurrent weights \citep{van2019blockrnn} of 8 blocks to allow for a large number of memory units without quadratic increase in parameters and FLOPs. The input to the GRU at each time step is a linear embedding of the sampled latent $z_t$, of the action $a_t$, and of the recurrent state to allow mixing between blocks.

\clearpage
\vspace*{-3\baselineskip}
\subsection*{Benchmarks}

\paragraph{Protocols}
Summarized in \cref{tab:benchmarks}, we follow the standard evaluation protocols for the benchmarks where established.
Atari \citep{bellemare2013ale} uses 57 tasks with sticky actions\citep{machado2018stickyactions}. The random and human reference scores used to normalize scores vary across the literature and we chose the most common reference values, replicated in \cref{tab:atari}.
DMLab\citep{beattie2016dmlab} uses 30 tasks\citep{espeholt2018impala} and we use the fixed action space\citep{hessel2019popart,kapturowski2018r2d2}. We evaluate at 100M steps because running for 10B as in some prior work was infeasible. Because existing published baselines perform poorly at 100M steps, we compare to their performance at 1B steps instead, giving them a $10\times$ data advantage.
ProcGen uses the hard difficulty setting and the unlimited level set \citep{cobbe2020procgen}. Prior work compares at different step budgets\citep{cobbe2020procgen,cobbe2021ppg} and we compare at 50M steps due to computational cost, as there is no action repeat.
For Minecraft Diamond purely from sparse rewards, we establish the evaluation protocol to report the episode return measured at 100M env steps, corresponding to about 100 days of in-game time.
Atari100k\citep{kaiser2019simple} includes 26 tasks with a budget of 400K env steps, 100K after action repeat. Prior work has used various environment settings, summarized in \cref{tab:atari100k_setting}, and we chose the environments as originally introduced.
Visual Control\citep{tassa2018dmcontrol,hafner2019dreamer} spans 20 tasks with an action repeat of 2.
Proprioceptive Control follows the same protocol but we exclude the two quadruped tasks because of baseline availability in prior work\citep{yarats2021drqv2}.

\begin{table}[h!]
\vspace*{-0.3ex}
\centering
\begin{mytabular}{
  colspec = {| L{7em} | C{3.3em} C{3.3em} C{3.3em} C{3.3em} C{3.3em} C{3.3em} C{3.3em} |},
  row{1} = {font=\bfseries},
  stretch=0.9,
}

\toprule
\newline Benchmark & \newline Tasks & Env Steps & Action Repeat & Env Instances & Replay Ratio & GPU Days & Model Size \\
\midrule
Minecraft       & \o1 &   100M & 1 &  64 & \o\o32 & \o8.9 & 200M \\
DMLab           &  30 &   100M & 4 &  16 & \o\o32 & \o2.9 & 200M \\
ProcGen         &  16 &  \o50M & 1 &  16 & \o\o64 &  16.1 & 200M \\
Atari           &  57 &   200M & 4 &  16 & \o\o32 & \o7.7 & 200M \\
Atari100K       &  26 &   400K & 4 & \o1 &  \o128 & \o0.1 & 200M \\
BSuite          &  23 &  \o--- & 1 & \o1 &   1024 & \o0.5 & 200M \\
Proprio Control &  18 &   500K & 2 &  16 &  \o512 & \o0.3 & \o12M \\
Visual Control  &  20 & \o\o1M & 2 &  16 &  \o512 & \o0.1 & \o12M \\
\bottomrule

\end{mytabular}
\caption{Benchmark overview. All agents were trained on a single Nvidia A100 GPU each.}
\label{tab:benchmarks}
\vspace*{-1.3ex}
\end{table}

\paragraph{Environment instances}
In earlier experiments, we observed that the performance of both Dreamer and PPO is robust to the number of environment instances. Based on the CPU resources available on our training machines, we use 16 environment instance by default. For BSuite, the benchmark requires using a single environment instance. We also use a single environment instance for Atari100K because the benchmark has a budget of 400K env steps whereas the maximum episode length in Atari is in principle 432K env steps. For Minecraft, we use 64 environments using remote CPU workers to speed up experiments because the environment is slower to step.

\paragraph{Seeds and error bars}
We run 5 seeds for each Dreamer and PPO per benchmark, with the exception of 1 seed for ProcGen due to computational constraints, 10 seeds for BSuite as required by the benchmark, and 10 seeds for Minecraft to reliably report the fraction of runs that achieve diamonds. All curves show the mean over seeds with one standard deviation shaded.

\paragraph{Computational choices}
All Dreamer and PPO agents in this paper were trained on a single Nvidia A100 GPU each. Dreamer uses the 200M model size by default. On the two control suitse, Dreamer the same performance using the substantially faster 12M model, making it more accessible to researchers. The replay ratio control the trade-off between computational cost and data efficiency as analyzed in \cref{fig:ablations} and is chosen to fit the step budget of each benchmark.

\clearpage

\subsection*{Model sizes}

To accommodate different computational budgets and analyze robustness to different model sizes, we define a range of models ranging from 12M to 400M parameters shown in \cref{tab:modelsizes}. The sizes are parameterized by the model dimension, which approximately increases in multiples of 1.5, alternating between powers of two and power of two scaled by 1.5. This yields tensor shapes that are multiples of 8 as required for hardware efficiency.
Sizes of different network components derive from the model dimension. The MLPs have the model dimension as the number of hidden units. The sequence model has 8 times the number of recurrent units, split into 8 blocks of the same size as the MLPs. The convolutional encoder and decoder layers closest to the data use $16\times$ fewer channels than the model dimension. Each latent also uses $16\times$ fewer codes than the model dimension. The number of hidden layers and number of latents is fixed across model sizes.
All hyperparamters, including the learning rate and batch size, are fixed across model sizes.

\begin{table}[h!]
\centering
\begin{mytabular}{
  colspec = {| L{12em} | C{3.5em} C{3.5em} C{3.5em} C{3.5em} C{3.5em} C{3.5em} |},
  row{1} = {font=\bfseries},
}

\toprule
Parameters                 &  12M  &   25M &   50M &  100M &  200M &     400M \\
\midrule
Hidden size ($d$)          & \o256 & \o384 & \o512 & \o768 &  1024 &   \o1536 \\
Recurrent units ($8d$)     &  1024 &  3072 &  4096 &  6144 &  8192 &    12288 \\
Base CNN channels ($d/16$) &\o\o16 &\o\o24 &\o\o32 &\o\o48 &\o\o64 & \o\o\o96 \\
Codes per latent ($d/16$)  &\o\o16 &\o\o24 &\o\o32 &\o\o48 &\o\o64 & \o\o\o96 \\
\bottomrule

\end{mytabular}
\caption{Dreamer model sizes. The number of MLP hidden units defines the model dimension, from which recurrent units, convolutional channels, and number of codes per latent are derived. The number of layers and latents is constant across model sizes.
}
\label{tab:modelsizes}
\end{table}

\subsection*{Previous Dreamer generations}

We present the third generation of the Dreamer line of work. Where the distinction is useful, we refer to this algorithm as DreamerV3. The DreamerV1 algorithm\citep{hafner2019dreamer} was limited to continuous control, the DreamerV2 algorithm\citep{hafner2020dreamerv2} surpassed human performance on Atari, and the DreamerV3 algorithm enables out-of-the-box learning across diverse benchmarks.

We summarize the changes that DreamerV3 introduces as follows:
\begin{itemize}
\item Robustness techniques: Observation symlog, KL balance and free bits, 1\% unimix for all categoricals, percentile return normalization, symexp twohot loss for the reward head and critic.
\item Network architecture: Block GRU, RMSNorm normalization, SiLu activation.
\item Optimizer: Adaptive gradient clipping (AGC), LaProp (RMSProp followed by momentum).
\item Replay buffer: Larger capacity, online queue, storing and updating latent states.
\end{itemize}

\clearpage
\subsection*{Hyperparameters}
\begin{table}[h!]
\centering
\begin{mytabular}{
  colspec = {| L{15em} | C{6em} | C{10em} |},
  row{1} = {font=\bfseries},
}

\toprule
\textbf{Name} & \textbf{Symbol} & \textbf{Value} \\
\midrule
\multicolumn{3}{l}{\textbf{General}} \\
\midrule
Replay capacity & --- & $5 \times 10^6\!\!$ \\
Batch size & $B$ & 16 \\
Batch length & $T$ & 64 \\
Activation & --- & $\operatorname{RMSNorm}+\operatorname{SiLU}$ \\
Learning rate & --- & $4 \times 10^{-5}$ \\
Gradient clipping & --- & $\operatorname{AGC}(0.3)$ \\
Optimizer & --- & $\operatorname{LaProp}(\epsilon=10^{-20})$ \\
\midrule
\multicolumn{3}{l}{\textbf{World Model}} \\
\midrule
Reconstruction loss scale & $\beta_{\mathrm{pred}}$ & 1 \\
Dynamics loss scale & $\beta_{\mathrm{dyn}}$ & 1 \\
Representation loss scale & $\beta_{\mathrm{rep}}$ & 0.1 \\
Latent unimix & --- & $1\%$ \\
Free nats & --- & 1 \\
\midrule
\multicolumn{3}{l}{\textbf{Actor Critic}} \\
\midrule
Imagination horizon & $H$ & 15 \\
Discount horizon & $1/(1-\gamma)$ & 333 \\
Return lambda & $\lambda$ & 0.95 \\
Critic loss scale & $\beta_{\mathrm{val}}$ & 1 \\
Critic replay loss scale & $\beta_{\mathrm{repval}}$ & 0.3 \\
Critic EMA regularizer & --- & 1 \\
Critic EMA decay & --- & 0.98 \\
Actor loss scale & $\beta_{\mathrm{pol}}$ & 1 \\
Actor entropy regularizer & $\eta$ & $3 \times 10^{-4}$ \\
Actor unimix & --- & $1\%$ \\
Actor RetNorm scale & $S$ & $\operatorname{Per}(R, 95) - \operatorname{Per}(R, 5)$ \\
Actor RetNorm limit & $L$ & 1 \\
Actor RetNorm decay & --- & 0.99 \\
\bottomrule

\end{mytabular}
\caption{Dreamer hyperparameters. The same values are used across all benchmarks, including proprioceptive and visual inputs, continuous and discrete actions, and 2D and 3D domains. We do not use any hyperparameter annealing, prioritized replay, weight decay, or dropout.}
\label{tab:hparams}
\end{table}

\clearpage
\subsection*{Minecraft}
\label{sec:mcenv}

\paragraph{Game description}
With 100M monthly active users, Minecraft is one of the most popular video games worldwide.
Minecraft features a procedurally generated 3D world of different biomes, including plains, forests, jungles, mountains, deserts, taiga, snowy tundra, ice spikes, swamps, savannahs, badlands, beaches, stone shores, rivers, and oceans.
The world consists of 1 meter sized blocks that the player and break and place.
There are about 30 different creatures that the player can interact with or fight.
From gathered resources, the player can use over 350 recipes to craft new items and progress through the technology tree, all while ensuring safety and food supply to survive.
There are many conceivable tasks in Minecraft and as a first step, the research community has focused on the salient task of obtaining a diamonds, a rare item found deep underground and requires progressing through the technology tree.

\paragraph{Learning environment}
We built the Minecraft Diamond environment on top of MineRL to define a flat categorical action space and fix issues we discovered with the original environments via human play testing.
For example, when breaking diamond ore, the item sometimes jumps into the inventory and sometimes needs to be collected from the ground.
The original environment terminates episodes when breaking diamond ore so that many successful episodes end before collecting the item and thus without the reward.
We remove this early termination condition and end episodes when the player dies or after 36000 steps, corresponding to 30 minutes at the control frequency of 20Hz.
Another issue is that the jump action has to be held for longer than one control step to trigger a jump, which we solve by keeping the key pressed in the background for 200ms.
We built the environment on top of MineRL v0.4.4 \citep{guss2019minerl}, which offers abstract crafting actions. The Minecraft version is 1.11.2.

\paragraph{Observations and rewards}
The agent observes a $64\times64\times3$ first-person image, an inventory count vector for the over 400 items, a vector of maximum inventory counts since episode begin to tell the agent which milestones it has achieved, a one-hot vector indicating the equipped item, and scalar inputs for the health, hunger, and breath levels.
We follow the sparse reward structure of the MineRL competition environment\citep{guss2019minerl} that rewards 12 milestones leading up to the diamond, for obtaining the items log, plank, stick, crafting table, wooden pickaxe, cobblestone, stone pickaxe, iron ore, furnace, iron ingot, iron pickaxe, and diamond.
The reward for each item is only given once per episode, and the agent has to learn to collect certain items multiple times to achieve the next milestone.
To make the return easy to interpret, we give a reward of $+1$ for each milestone instead of scaling rewards based on how valuable each item is.
Additionally, we give $-0.01$ for each lost heart and $+0.01$ for each restored heart, but did not investigate whether this is helpful.

\clearpage
\section*{Supplementary material}

\subsection*{Minecraft video predictions}
\begin{figure}[ht!]
\centering
\includegraphics[width=\linewidth,trim={0 .5cm 0 0},clip]{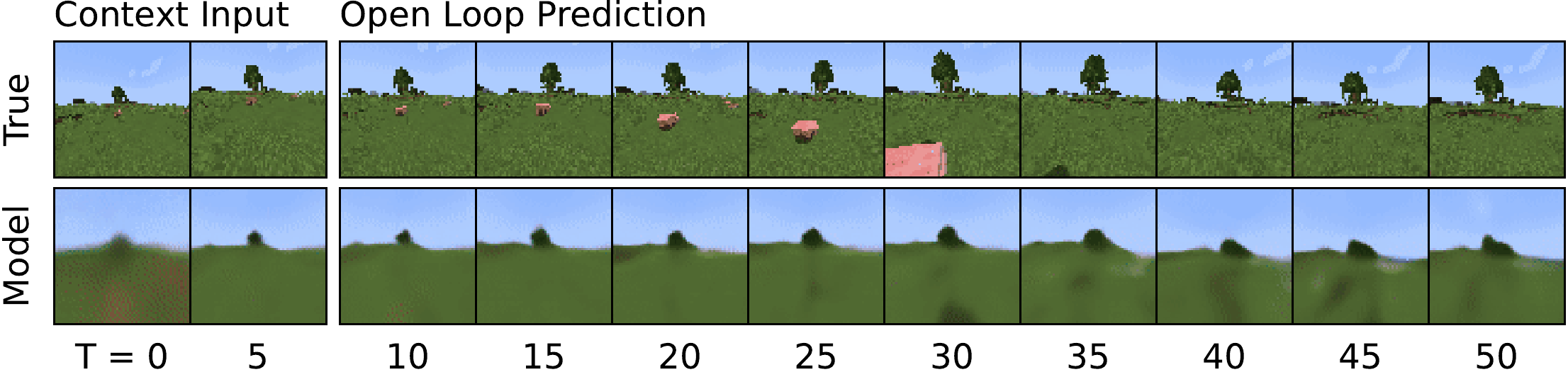} \\[1ex]
\includegraphics[width=\linewidth,trim={0 .5cm 0 .5cm},clip]{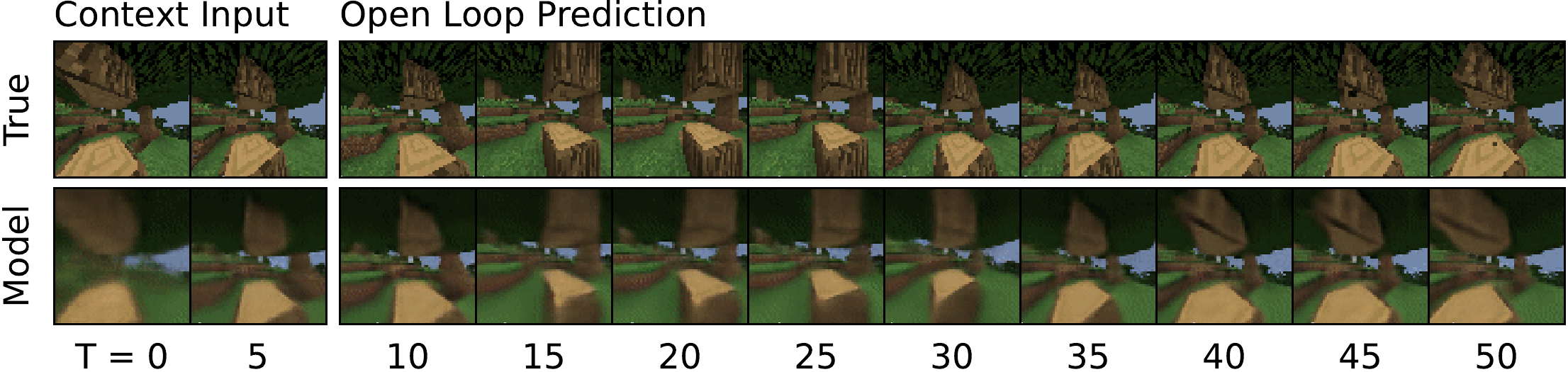} \\[1ex]
\includegraphics[width=\linewidth,trim={0 .5cm 0 .5cm},clip]{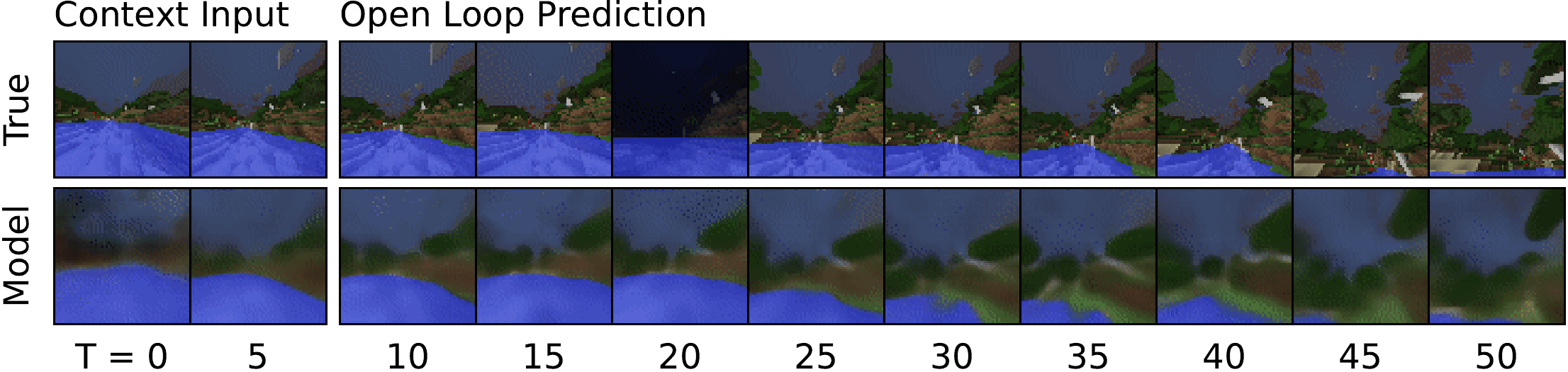} \\[1ex]
\includegraphics[width=\linewidth,trim={0 .5cm 0 .5cm},clip]{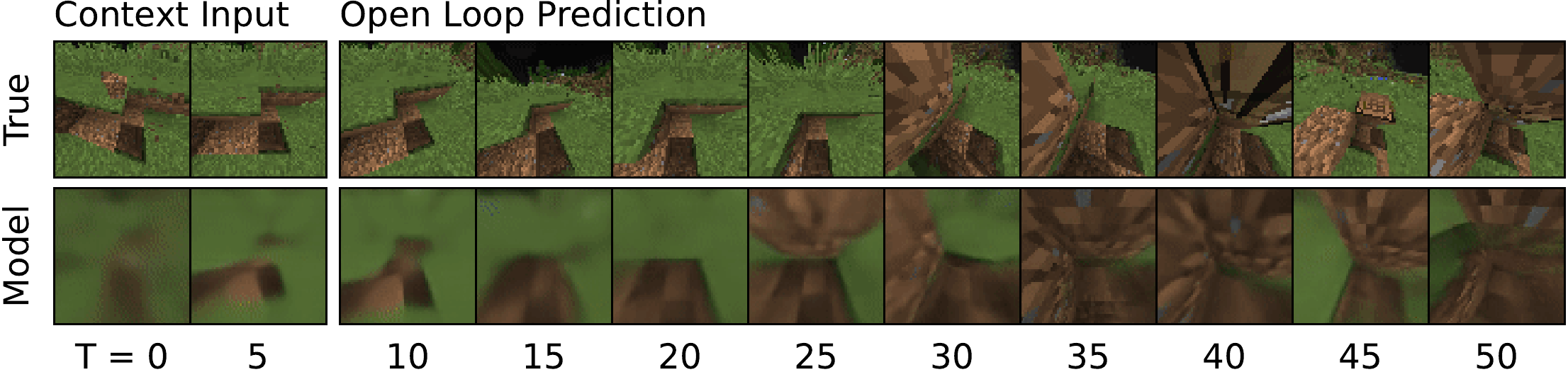} \\[1ex]
\includegraphics[width=\linewidth,trim={0 0 0 .5cm},clip]{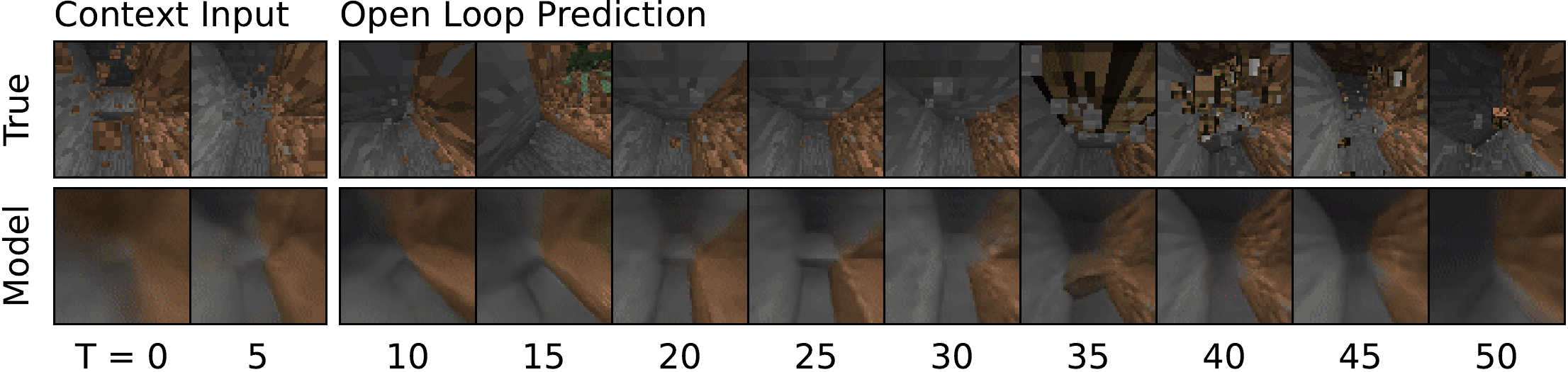} \\[1ex]
\caption{Multi-step predictions on Minecraft. The world model receives the first 5 frames as context input and the predicts 45 steps into the future given the action sequence and without access to intermediate images.}
\label{fig:openl_more}
\end{figure}

\vspace*{-4ex}

\clearpage
\subsection*{Minecraft additional results}
\noindent
\begin{minipage}[b]{.46\textwidth}
\centering
\begin{mytabular}{
  colspec = {| L{6em} | C{6em} |},
  row{1} = {font=\bfseries},
}

\toprule
Method & Return \\
\midrule
Dreamer & 9.1 \\
IMPALA & 7.1 \\
Rainbow & 6.3 \\
PPO & 5.1 \\
\bottomrule

\end{mytabular}
\captionof{table}{Minecraft Diamond scores at 100M environment steps.}
\label{tab:minecraft}

\end{minipage}\hfill
\begin{minipage}[b]{.46\textwidth}
\centering
\includegraphics[width=\linewidth]{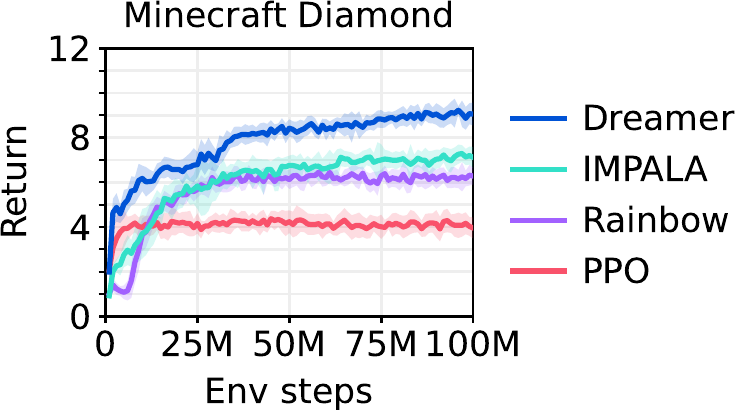}
\captionof{figure}{Minecraft learning curves.}
\label{fig:mcreturn}

\end{minipage}

\begin{figure}[h!]
\centering
\vspace*{3ex}
\includegraphics[width=\linewidth]{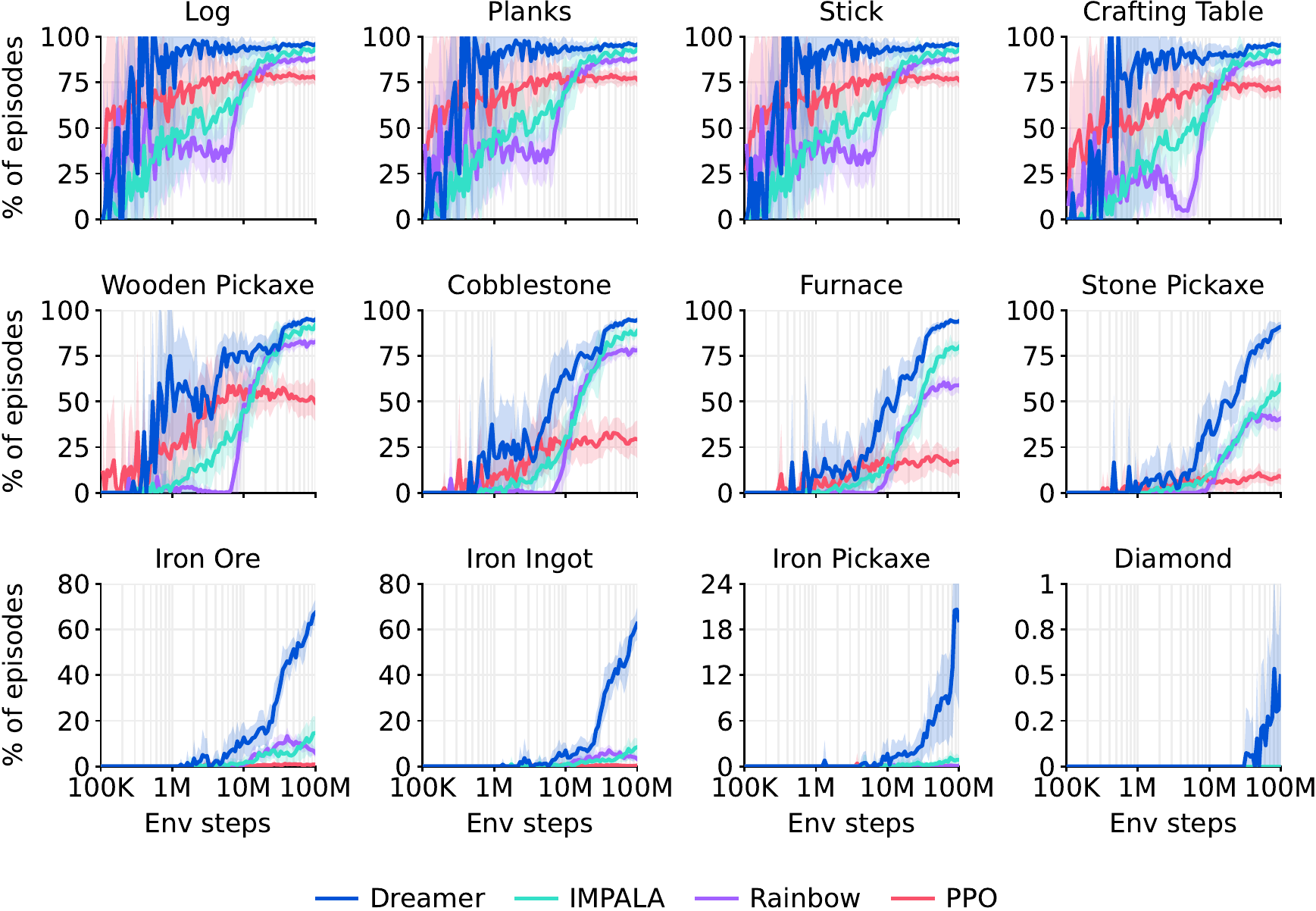}
\caption{Item success rates as a percentage of episodes. Dreamer obtains items at substantially higher rates than the baselines and continues to improve until the 100M step budget.
At the budget, Dreamer obtains diamonds in 0.4\% of episodes, leaving a challenge for future research. This metric differs from \cref{fig:minecraft}, which shows that over the course of training, 100\% of Dreamer agents obtain one or more diamonds regardless of episode boundaries, compared to 0\% of the baseline agents.}
\label{fig:mcitems}
\end{figure}

\clearpage
\subsection*{Atari learning curves}
\begin{figure}[h!]
\vspace*{-1ex}
\centering
\begin{adjustwidth}{0ex}{-1ex}
\includegraphics[width=\linewidth]{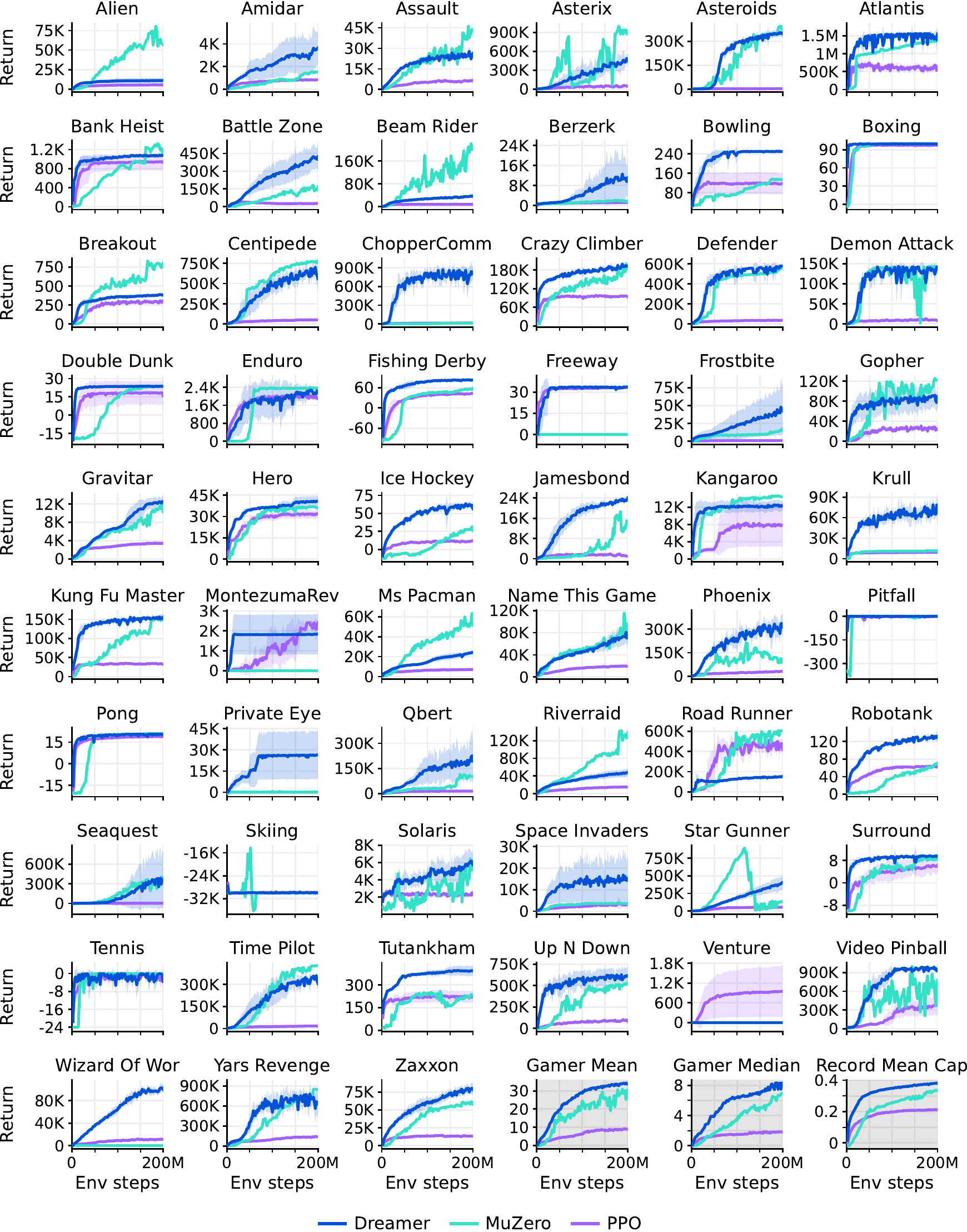}
\end{adjustwidth}
\caption{Atari learning curves.}
\label{fig:atari}
\vspace*{-1ex}
\end{figure}

\clearpage
\subsection*{Atari scores}
\vspace*{-1ex}
\begin{table}[h!]
\centering\small
\begin{mytabular}{
  colspec = {| L{10.3em} | C{4em} C{4em} C{4em} | C{4.4em} C{4.4em} C{4.4em} |},
  row{1} = {font=\bfseries},
  stretch = 0.4,
}
\toprule
Task & Random & Gamer & Record & PPO & MuZero & Dreamer \\
\midrule
Environment steps & --- & --- & --- & \o200M & \o\o200M & \o\o200M \\
\midrule

Alien & \o\o228 & \o7128 & \o\o251916 & \o\o5476 & \textbf{\o\o56835} & \o\o10977 \\
Amidar & \o\o\o\o6 & \o1720 & \o\o104159 & \o\o\o817 & \o\o\o1517 & \textbf{\o\o\o3612} \\
Assault & \o\o222 & \o\o742 & \o\o\o\o8647 & \o\o6673 & \textbf{\o\o42742} & \o\o26010 \\
Asterix & \o\o210 & \o8503 & \o1000000 & \o47190 & \textbf{\o879375} & \o441763 \\
Asteroids & \o\o719 & 47389 & 10506650 & \o\o2479 & \textbf{\o374146} & \o348684 \\
Atlantis & 12850 & 29028 & 10604840 & 539721 & 1353617 & \textbf{1553222} \\
Bank Heist & \o\o\o14 & \o\o753 & \o\o\o82058 & \o\o\o946 & \textbf{\o\o\o1077} & \textbf{\o\o\o1083} \\
Battle Zone & \o2360 & 37188 & \o\o801000 & \o27816 & \o167412 & \textbf{\o419653} \\
Beam Rider & \o\o364 & 16926 & \o\o999999 & \o\o7973 & \textbf{\o201154} & \o\o37073 \\
Berzerk & \o\o124 & \o2630 & \o1057940 & \o\o1186 & \o\o\o1698 & \textbf{\o\o10557} \\
Bowling & \o\o\o23 & \o\o161 & \o\o\o\o\o300 & \o\o\o118 & \o\o\o\o133 & \textbf{\o\o\o\o250} \\
Boxing & \o\o\o\o0 & \o\o\o12 & \o\o\o\o\o100 & \textbf{\o\o\o\o98} & \textbf{\o\o\o\o100} & \textbf{\o\o\o\o100} \\
Breakout & \o\o\o\o2 & \o\o\o30 & \o\o\o\o\o864 & \o\o\o299 & \textbf{\o\o\o\o799} & \o\o\o\o384 \\
Centipede & \o2091 & 12017 & \o1301709 & \o51833 & \textbf{\o774421} & \o554553 \\
Chopper Command & \o\o811 & \o7388 & \o\o999999 & \o12667 & \o\o\o8945 & \textbf{\o802698} \\
Crazy Climber & 10780 & 35829 & \o\o219900 & \o93176 & \textbf{\o184394} & \textbf{\o193204} \\
Defender & \o2874 & 18689 & \o6010500 & \o38270 & \textbf{\o554492} & \textbf{\o579875} \\
Demon Attack & \o\o152 & \o1971 & \o1556345 & \o\o8229 & \textbf{\o142509} & \textbf{\o142109} \\
Double Dunk & \o\o\o\llap{--}19 & \o\o\o\llap{--}16 & \o\o\o\o\o\o22 & \o\o\o\o16 & \textbf{\o\o\o\o\o23} & \textbf{\o\o\o\o\o24} \\
Enduro & \o\o\o\o0 & \o\o860 & \o\o\o\o9500 & \o\o1887 & \textbf{\o\o\o2369} & \o\o\o2166 \\
Fishing Derby & \o\o\o\llap{--}92 & \o\o\o\llap{--}39 & \o\o\o\o\o\o71 & \o\o\o\o43 & \o\o\o\o\o58 & \textbf{\o\o\o\o\o82} \\
Freeway & \o\o\o\o0 & \o\o\o30 & \o\o\o\o\o\o38 & \textbf{\o\o\o\o33} & \o\o\o\o\o\o0 & \textbf{\o\o\o\o\o34} \\
Frostbite & \o\o\o65 & \o4335 & \o\o454830 & \o\o1123 & \o\o17087 & \textbf{\o\o41888} \\
Gopher & \o\o258 & \o2412 & \o\o355040 & \o24792 & \textbf{\o122025} & \o\o87600 \\
Gravitar & \o\o173 & \o3351 & \o\o162850 & \o\o3436 & \o\o10301 & \textbf{\o\o12570} \\
Hero & \o1027 & 30826 & \o1000000 & \o31967 & \o\o36063 & \textbf{\o\o40677} \\
Ice Hockey & \o\o\o\llap{--}11 & \o\o\o\o1 & \o\o\o\o\o\o36 & \o\o\o\o12 & \o\o\o\o\o26 & \textbf{\o\o\o\o\o57} \\
Jamesbond & \o\o\o29 & \o\o303 & \o\o\o45550 & \o\o1019 & \o\o14872 & \textbf{\o\o24010} \\
Kangaroo & \o\o\o52 & \o3035 & \o1424600 & \o\o7769 & \textbf{\o\o14380} & \o\o12229 \\
Krull & \o1598 & \o2666 & \o\o104100 & \o\o9193 & \o\o11476 & \textbf{\o\o69858} \\
Kung Fu Master & \o\o258 & 22736 & \o1000000 & \o32335 & \textbf{\o148936} & \textbf{\o154893} \\
Montezuma Revenge & \o\o\o\o0 & \o4753 & \o1219200 & \textbf{\o\o2368} & \o\o\o\o\o\o0 & \o\o\o1852 \\
Ms Pacman & \o\o307 & \o6952 & \o\o290090 & \o\o7041 & \textbf{\o\o51310} & \o\o24079 \\
Name This Game & \o2292 & \o8049 & \o\o\o25220 & \o19441 & \textbf{\o\o85331} & \o\o77809 \\
Phoenix & \o\o761 & \o7243 & \o4014440 & \o31412 & \o105593 & \textbf{\o316606} \\
Pitfall & \o\o\llap{--}229 & \o6464 & \o\o114000 & \o\o\o\o\o\llap{--}2 & \textbf{\o\o\o\o\o\o0} & \textbf{\o\o\o\o\o\o0} \\
Pong & \o\o\o\llap{--}21 & \o\o\o15 & \o\o\o\o\o\o21 & \o\o\o\o19 & \textbf{\o\o\o\o\o21} & \textbf{\o\o\o\o\o20} \\
Private Eye & \o\o\o25 & 69571 & \o\o101800 & \o\o\o\o73 & \o\o\o\o100 & \textbf{\o\o26432} \\
Qbert & \o\o164 & 13455 & \o2400000 & \o14554 & \o102129 & \textbf{\o201084} \\
Riverraid & \o1338 & 17118 & \o1000000 & \o14860 & \textbf{\o137983} & \o\o48080 \\
Road Runner & \o\o\o12 & \o7845 & \o2038100 & 423995 & \textbf{\o604083} & \o150402 \\
Robotank & \o\o\o\o2 & \o\o\o12 & \o\o\o\o\o\o76 & \o\o\o\o63 & \o\o\o\o\o70 & \textbf{\o\o\o\o132} \\
Seaquest & \o\o\o68 & 42055 & \o\o999999 & \o\o1927 & \textbf{\o399764} & \o356584 \\
Skiing & \llap{--}17098 & \o\llap{--}4337 & \o\o\o\o\llap{--}3272 & \textbf{\o\llap{--}29926} & \textbf{\o\o\llap{--}30000} & \textbf{\o\o\llap{--}29965} \\
Solaris & \o1236 & 12327 & \o\o111420 & \o\o2368 & \textbf{\o\o\o5860} & \textbf{\o\o\o5851} \\
Space Invaders & \o\o148 & \o1669 & \o\o621535 & \o\o3489 & \o\o\o3639 & \textbf{\o\o15005} \\
Star Gunner & \o\o664 & 10250 & \o\o\o77400 & \o53439 & \o127417 & \textbf{\o408961} \\
Surround & \o\o\o\llap{--}10 & \o\o\o\o6 & \o\o\o\o\o\o\o6 & \o\o\o\o\o6 & \textbf{\o\o\o\o\o\o9} & \textbf{\o\o\o\o\o\o9} \\
Tennis & \o\o\o\llap{--}24 & \o\o\o\o\llap{--}8 & \o\o\o\o\o\o21 & \o\o\o\o\o\llap{--}1 & \textbf{\o\o\o\o\o\o0} & \o\o\o\o\o\o\llap{--}3 \\
Time Pilot & \o3568 & \o5229 & \o\o\o65300 & \o17250 & \textbf{\o427209} & \o314947 \\
Tutankham & \o\o\o11 & \o\o168 & \o\o\o\o5384 & \o\o\o225 & \o\o\o\o235 & \textbf{\o\o\o\o395} \\
Up N Down & \o\o533 & 11693 & \o\o\o82840 & \o83743 & \o522962 & \textbf{\o614065} \\
Venture & \o\o\o\o0 & \o1188 & \o\o\o38900 & \textbf{\o\o\o953} & \o\o\o\o\o\o0 & \o\o\o\o\o\o0 \\
Video Pinball & 16257 & 17668 & 89218328 & 382306 & \o775304 & \textbf{\o940631} \\
Wizard Of Wor & \o\o564 & \o4756 & \o\o395300 & \o10910 & \o\o\o\o\o\o0 & \textbf{\o\o99136} \\
Yars Revenge & \o3093 & 54577 & 15000105 & 137164 & \textbf{\o846061} & \o675774 \\
Zaxxon & \o\o\o32 & \o9173 & \o\o\o83700 & \o13599 & \o\o58115 & \textbf{\o\o78443} \\

\midrule

Gamer median (\%) & \o\o\o\o0 & \o\o100 & \o\o\o\o3716 & \o\o\o180 & \o\o\o\o693 & \textbf{\o\o\o\o830} \\
Gamer mean (\%) & \o\o\o\o0 & \o\o100 & \o\o123001 & \o\o\o892 & \o\o\o3054 & \textbf{\o\o\o3381} \\
Record mean (\%) & \o\o\o\o0 & \o\o\o13 & \o\o\o\o\o100 & \o\o\o\o21 & \o\o\o\o\o66 & \textbf{\o\o\o\o\o74} \\
Record mean capped \rlap{(\%)} & \o\o\o\o0 & \o\o\o13 & \o\o\o\o\o100 & \o\o\o\o21 & \o\o\o\o\o34 & \textbf{\o\o\o\o\o38} \\

\bottomrule
\end{mytabular}
\vspace*{-1ex}
\caption{Atari scores.}
\label{tab:atari}
\vspace*{-3ex}
\end{table}

\clearpage
\subsection*{ProcGen learning curves}
\begin{figure}[h!]
\centering
\includegraphics[width=\linewidth]{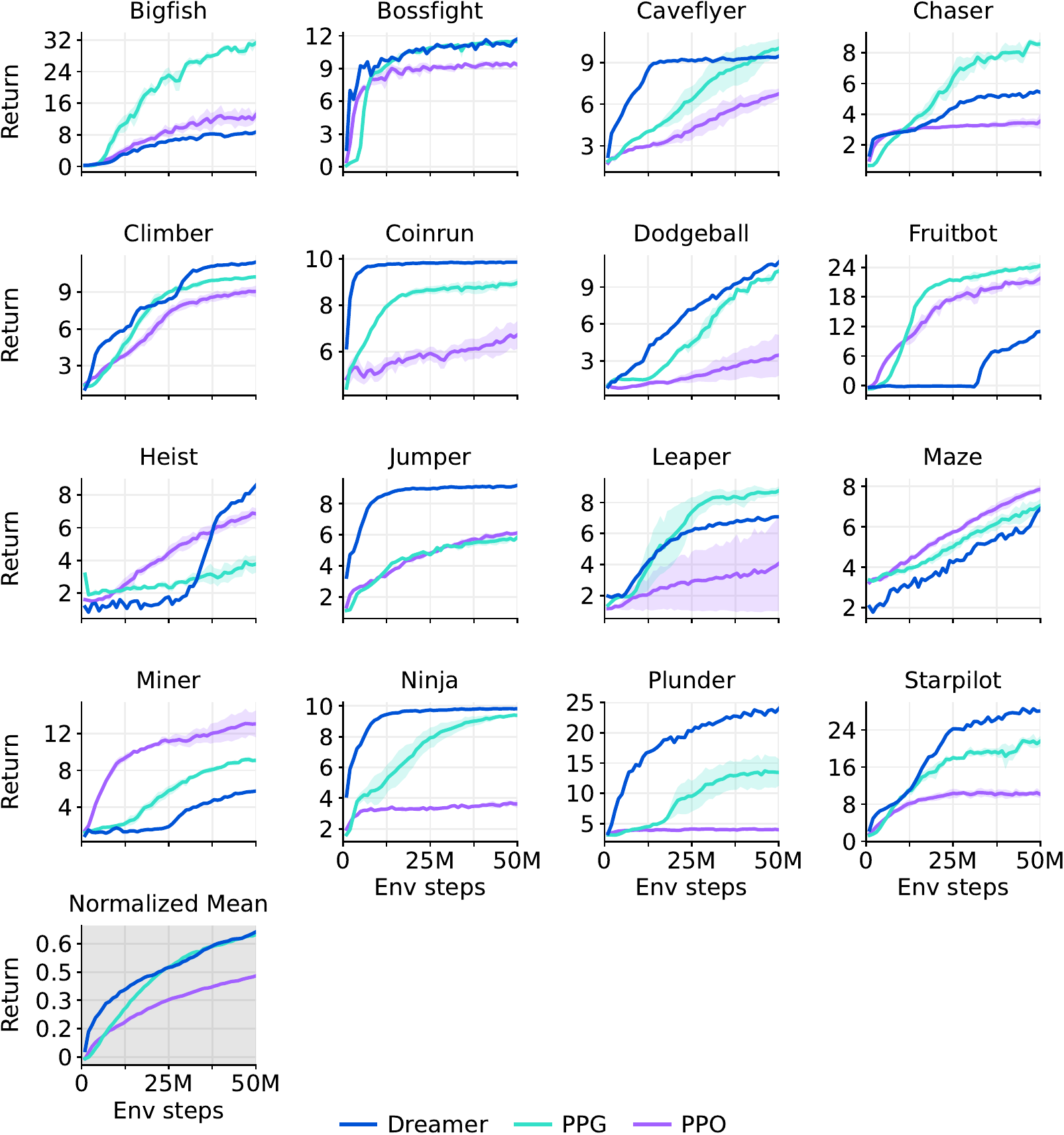}
\caption{ProcGen learning curves.}
\label{fig:procgen}
\end{figure}

\clearpage
\subsection*{ProcGen scores}
\begin{table*}[h!]
\centering
\begin{mytabular}{
  colspec = {| L{12em} | C{8em} C{5em} C{5em} C{5em} |},
  row{1} = {font=\bfseries},
}

\toprule
Task & Original PPO & PPO & PPG & Dreamer \\
\midrule
Environment steps & 50M & 50M & 50M & 50M \\
\midrule

Bigfish & 10.92 & 12.72 & \textbf{31.26} & \o8.62 \\
Bossfight & 10.47 & \o9.36 & \textbf{11.46} & \textbf{11.61} \\
Caveflyer & \o6.03 & \o6.71 & \textbf{10.02} & \o9.42 \\
Chaser & \o4.48 & \o3.54 & \textbf{\o8.57} & \o5.49 \\
Climber & \o7.59 & \o9.04 & 10.24 & \textbf{11.43} \\
Coinrun & \o7.93 & \o6.71 & \o8.98 & \textbf{\o9.86} \\
Dodgeball & \o4.80 & \o3.44 & 10.31 & \textbf{10.93} \\
Fruitbot & 20.28 & 21.69 & \textbf{24.32} & 11.04 \\
Heist & \o2.25 & \o6.87 & \o3.77 & \textbf{\o8.51} \\
Jumper & \o5.09 & \o6.13 & \o5.84 & \textbf{\o9.17} \\
Leaper & \o5.90 & \o4.07 & \textbf{\o8.76} & \o7.05 \\
Maze & \o4.97 & \textbf{\o7.86} & \o7.06 & \o6.85 \\
Miner & \o7.56 & \textbf{12.97} & \o9.08 & \o5.71 \\
Ninja & \o6.16 & \o3.62 & \textbf{\o9.38} & \textbf{\o9.82} \\
Plunder & 11.16 & \o3.99 & 13.44 & \textbf{23.81} \\
Starpilot & 17.00 & 10.13 & 21.57 & \textbf{28.00} \\
\midrule
Normalized mean & 41.16 & 42.80 & \textbf{64.89} & \textbf{66.01} \\

\bottomrule
\end{mytabular}
\caption{ProcGen scores. The PPO implementation we use throughout our paper under fixed hyperparameters performs on par or better than the original PPO, which its authors describe as highly tuned with near optimal hyperparameters \citep{cobbe2021ppg}.}
\label{tab:procgen}
\end{table*}

\clearpage
\subsection*{DMLab learning curves}
\begin{figure}[h!]
\centering
\includegraphics[width=\linewidth]{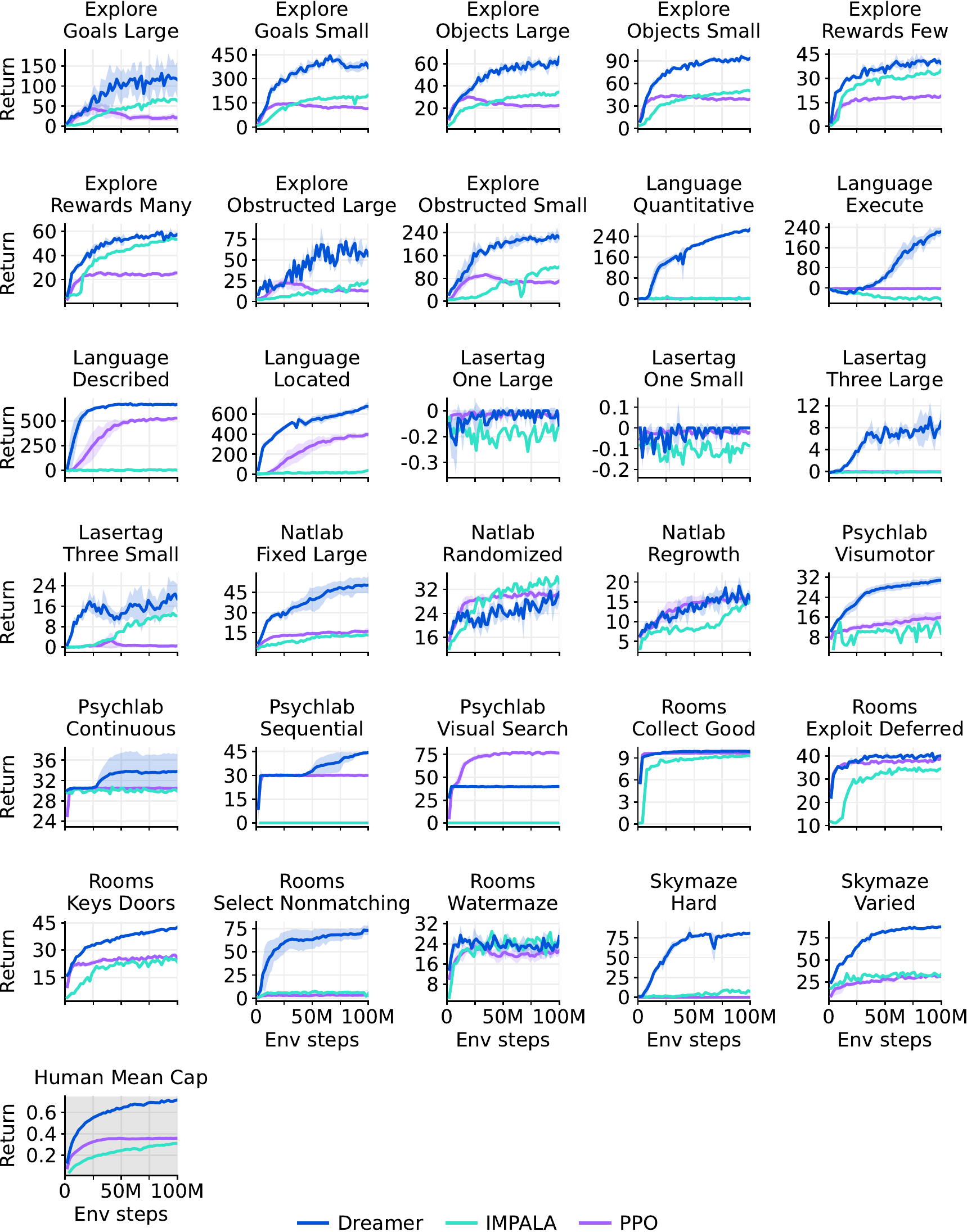}
\caption{DMLab learning curves.}
\label{fig:dmlab}
\vspace*{-3ex}
\end{figure}

\clearpage
\subsection*{DMLab scores}
\begin{table}[h!]
\centering
\footnotesize
\begin{mytabular}{
      colspec = {| L{17em} | C{3.5em} C{3.7em} C{3.7em} | C{3.7em} C{3.4em} C{4em} |},
  row{1} = {font=\bfseries},
  stretch = 1.0,
}

\toprule
Task & R2D2+ & \clap{IMPALA} & \clap{IMPALA} & \clap{IMPALA} & PPO & Dreamer \\
\midrule
Environment steps & 10B & 10B & 1B & 100M & 100M & 100M \\
\midrule
Explore Goal Locations Large & 174.7 & 316.0 & 137.8 & \o64.2 & \o21.2 & \textbf{116.7} \\
Explore Goal Locations Small & 460.7 & 482.0 & 302.8 & 196.1 & 115.1 & \textbf{372.8} \\
Explore Object Locations Large & \o60.6 & \o91.0 & \o55.1 & \o34.3 & \o22.5 & \textbf{\o63.9} \\
Explore Object Locations Small & \o83.7 & 100.4 & \o75.9 & \o50.6 & \o38.9 & \textbf{\o93.5} \\
Explore Object Rewards Few & \o80.7 & \o92.6 & \o46.9 & \o34.6 & \o19.0 & \textbf{\o40.0} \\
Explore Object Rewards Many & \o75.8 & \o89.4 & \o68.5 & \o53.3 & \o25.5 & \textbf{\o58.1} \\
Explore Obstructed Goals Large & \o95.5 & 102.0 & \o57.9 & \o23.7 & \o12.9 & \textbf{\o52.8} \\
Explore Obstructed Goals Small & 311.9 & 372.0 & 214.9 & 118.0 & \o70.4 & \textbf{224.6} \\
Language Answer Quantitative Question & 344.4 & 362.0 & 304.7 & \o\o1.0 & \o\o0.3 & \textbf{266.0} \\
Language Execute Random Task & 497.4 & 465.4 & 140.8 & \o\llap{--}44.4 & \o\o\llap{--}2.5 & \textbf{223.7} \\
Language Select Described Object & 617.6 & 664.0 & 618.4 & \o\o0.2 & 526.9 & \textbf{665.5} \\
Language Select Located Object & 772.8 & 731.4 & 413.0 & \o38.2 & 397.5 & \textbf{679.5} \\
Lasertag One Opponent Large & \o\o0.0 & \o\o0.0 & \o\o0.0 & \o\o\llap{--}0.1 & \textbf{\o\o0.0} & \o\o\llap{--}0.1 \\
Lasertag One Opponent Small & \o31.8 & \o\o0.0 & \o\o0.0 & \o\o\llap{--}0.1 & \textbf{\o\o0.0} & \textbf{\o\o0.0} \\
Lasertag Three Opponents Large & \o28.6 & \o32.2 & \o10.4 & \o\o\llap{--}0.1 & \o\o0.0 & \textbf{\o\o9.0} \\
Lasertag Three Opponents Small & \o49.0 & \o57.2 & \o37.1 & \o12.1 & \o\o0.4 & \textbf{\o18.8} \\
Natlab Fixed Large Map & \o60.6 & \o63.4 & \o53.8 & \o13.0 & \o15.8 & \textbf{\o50.5} \\
Natlab Varying Map Randomized & \o42.4 & \o47.0 & \o40.5 & \textbf{\o35.3} & \o29.6 & \o31.2 \\
Natlab Varying Map Regrowth & \o24.6 & \o34.0 & \o25.5 & \o15.3 & \textbf{\o16.3} & \textbf{\o16.7} \\
Psychlab Arbitrary Visuomotor Mapping & \o33.1 & \o38.4 & \o16.5 & \o11.9 & \o16.0 & \textbf{\o30.7} \\
Psychlab Continuous Recognition & \o30.0 & \o28.6 & \o30.0 & \o30.1 & \o30.5 & \textbf{\o33.8} \\
Psychlab Sequential Comparison & \o30.0 & \o29.6 & \o\o0.0 & \o\o0.0 & \o30.0 & \textbf{\o44.3} \\
Psychlab Visual Search & \o79.9 & \o80.0 & \o\o0.0 & \o\o0.0 & \textbf{\o76.6} & \o40.1 \\
Rooms Collect Good Objects Test & \o\o9.9 & \o10.0 & \o\o9.9 & \o\o9.3 & \textbf{\o\o9.7} & \textbf{\o\o9.9} \\
Rooms Exploit Deferred Effects Test & \o38.1 & \o62.2 & \o37.6 & \o34.5 & \textbf{\o39.0} & \textbf{\o40.4} \\
Rooms Keys Doors Puzzle & \o46.2 & \o54.6 & \o36.9 & \o24.2 & \o26.0 & \textbf{\o42.4} \\
Rooms Select Nonmatching Object & \o63.6 & \o39.0 & \o63.2 & \o\o4.0 & \o\o2.7 & \textbf{\o73.1} \\
Rooms Watermaze & \o49.0 & \o47.0 & \o50.1 & \o23.6 & \o21.2 & \textbf{\o26.1} \\
Skymaze Irreversible Path Hard & \o76.0 & \o80.0 & \o46.4 & \o\o7.7 & \o\o0.0 & \textbf{\o80.9} \\
Skymaze Irreversible Path Varied & \o76.0 & 100.0 & \o69.8 & \o32.7 & \o31.3 & \textbf{\o87.7} \\
\midrule
Human mean capped (\%) & \o85.4 & \o85.1 & \o66.3 & \o31.0 & \o35.9 & \textbf{\o71.4} \\
\bottomrule

\end{mytabular}
\vspace{-0.5ex}
\caption{DMLab scores at 100M environment steps and larger budgets. The IMPALA agent corresponds to ``IMPALA (deep)'' presented by \citet{kapturowski2018r2d2} who made the learning curves available.}
\label{tab:dmlab}
\end{table}

\clearpage
\subsection*{Atari100k learning curves}
\begin{figure}[h!]
\centering
\includegraphics[width=\linewidth]{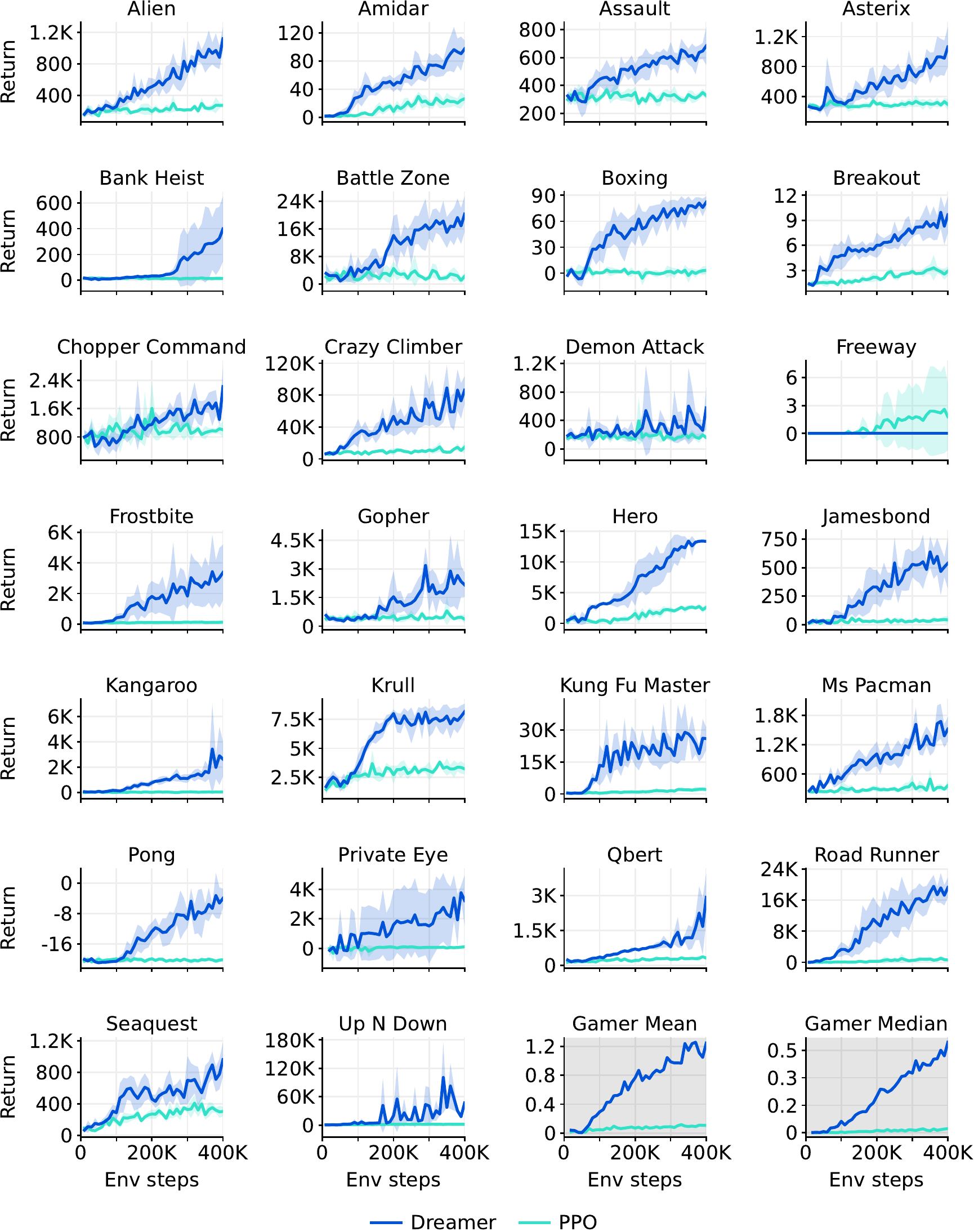}
\caption{Atari100k learning curves.}
\label{fig:atari100k}
\end{figure}

\clearpage
\subsection*{Atari100k scores}
\begin{table}[h!]
\vspace*{-1ex}
\centering
\small
\begin{mytabular}{
  colspec = {| L{8em} | C{3em} C{3em} | C{3em} C{3em} C{3em} C{3em} C{3em} C{4em} |},
  row{1} = {font=\bfseries},
  stretch = 0.6,
}

\toprule
Task & Random & Human & PPO & SimPLe & SPR & TWM & IRIS & Dreamer \\
\midrule
Environment steps & --- & --- & 400K & 400K & 400K & 400K & 400K & 400K \\

\midrule
Alien & \o\o228 & \o7128 & \o\o276 & \o\o617 & \o\o842 & \o\o675 & \o\o420 & \textbf{\o1118} \\
Amidar & \o\o\o\o6 & \o1720 & \o\o\o26 & \o\o\o74 & \textbf{\o\o180} & \o\o122 & \o\o143 & \o\o\o97 \\
Assault & \o\o222 & \o\o742 & \o\o327 & \o\o527 & \o\o566 & \o\o683 & \textbf{\o1524} & \o\o683 \\
Asterix & \o\o210 & \o8503 & \o\o292 & \textbf{\o1128} & \o\o962 & \textbf{\o1117} & \o\o854 & \o1062 \\
Bank Heist & \o\o\o14 & \o\o753 & \o\o\o14 & \o\o\o34 & \o\o345 & \textbf{\o\o467} & \o\o\o53 & \o\o398 \\
Battle Zone & \o2360 & 37188 & \o2233 & \o4031 & 14834 & \o5068 & 13074 & \textbf{20300} \\
Boxing & \o\o\o\o0 & \o\o\o12 & \o\o\o\o3 & \o\o\o\o8 & \o\o\o36 & \textbf{\o\o\o78} & \o\o\o70 & \textbf{\o\o\o82} \\
Breakout & \o\o\o\o2 & \o\o\o30 & \o\o\o\o3 & \o\o\o16 & \o\o\o20 & \o\o\o20 & \textbf{\o\o\o84} & \o\o\o10 \\
Chopper Command & \o\o811 & \o7388 & \o1005 & \o\o979 & \o\o946 & \o1697 & \o1565 & \textbf{\o2222} \\
Crazy Climber & 10780 & 35829 & 14675 & 62584 & 36700 & 71820 & 59324 & \textbf{86225} \\
Demon Attack & \o\o152 & \o1971 & \o\o160 & \o\o208 & \o\o518 & \o\o350 & \textbf{\o2034} & \o\o577 \\
Freeway & \o\o\o\o0 & \o\o\o30 & \o\o\o\o2 & \o\o\o17 & \o\o\o19 & \o\o\o24 & \textbf{\o\o\o31} & \o\o\o\o0 \\
Frostbite & \o\o\o65 & \o4335 & \o\o127 & \o\o237 & \o1171 & \o1476 & \o\o259 & \textbf{\o3377} \\
Gopher & \o\o258 & \o2412 & \o\o368 & \o\o597 & \o\o661 & \o1675 & \textbf{\o2236} & \textbf{\o2160} \\
Hero & \o1027 & 30826 & \o2596 & \o2657 & \o5859 & \o7254 & \o7037 & \textbf{13354} \\
Jamesbond & \o\o\o29 & \o\o303 & \o\o\o41 & \o\o100 & \o\o366 & \o\o362 & \o\o463 & \textbf{\o\o540} \\
Kangaroo & \o\o\o52 & \o3035 & \o\o\o55 & \o\o\o51 & \textbf{\o3617} & \o1240 & \o\o838 & \o2643 \\
Krull & \o1598 & \o2666 & \o3222 & \o2205 & \o3682 & \o6349 & \o6616 & \textbf{\o8171} \\
Kung Fu Master & \o\o258 & 22736 & \o2090 & 14862 & 14783 & 24555 & 21760 & \textbf{25900} \\
Ms Pacman & \o\o307 & \o6952 & \o\o366 & \o1480 & \o1318 & \textbf{\o1588} & \o\o999 & \textbf{\o1521} \\
Pong & \o\o\o\llap{--}21 & \o\o\o15 & \o\o\o\llap{--}20 & \o\o\o13 & \o\o\o\o\llap{--}5 & \textbf{\o\o\o19} & \o\o\o15 & \o\o\o\o\llap{--}4 \\
Private Eye & \o\o\o25 & 69571 & \o\o100 & \o\o\o35 & \o\o\o86 & \o\o\o87 & \o\o100 & \textbf{\o3238} \\
Qbert & \o\o164 & 13455 & \o\o317 & \o1289 & \o\o866 & \textbf{\o3331} & \o\o746 & \o2921 \\
Road Runner & \o\o\o12 & \o7845 & \o\o602 & \o5641 & 12213 & \o9109 & \o9615 & \textbf{19230} \\
Seaquest & \o\o\o68 & 42055 & \o\o305 & \o\o683 & \o\o558 & \o\o774 & \o\o661 & \textbf{\o\o962} \\
Up N Down & \o\o533 & 11693 & \o1502 & \o3350 & 10859 & 15982 & \o3546 & \textbf{46910} \\
\midrule
Gamer mean (\%) & \o\o\o\o0 & \o\o100 & \o\o\o11 & \o\o\o33 & \o\o\o62 & \o\o\o96 & \o\o105 & \textbf{\o\o125} \\
Gamer median (\%) & \o\o\o\o0 & \o\o100 & \o\o\o\o2 & \o\o\o13 & \o\o\o40 & \textbf{\o\o\o51} & \o\o\o29 & \textbf{\o\o\o49} \\
\bottomrule

\end{mytabular}
\vspace{-0.5ex}
\caption{Atari100k scores at 400K environment steps, corresponding to 100k agent steps.}
\label{tab:atari100k}
\end{table}

\begin{table}[h!]
\vspace*{-0.5ex}
\centering
\small
\begin{mytabular}{
  colspec = {| L{9em} | C{4.3em} C{4.3em} C{4.3em} C{4.3em} C{4.3em} C{4.3em} |},
  row{1} = {font=\bfseries},
  stretch=0.8,
}

\toprule
Setting & SimPLe & \clap{EffMuZero} & SPR & IRIS & TWM & Dreamer \\
\midrule
Gamer score (\%) & 33 & 190 & 62 & 105 & 96 & 125 \\
Gamer median (\%) & 13 & 109 & 40 & 29 & 51 & 49 \\
GPU days     & 5.0 & 0.6 & 0.1 & 3.5 & 0.4 & 0.1 \\
\midrule
Online planning        & --- &  X  & --- & --- & --- & --- \\
Data augmentation      & --- & --- &  X  & --- & --- & --- \\
Non-uniform replay     & --- &  X  &  X  & --- &  X  & --- \\
Separate hparams       & --- & --- & --- &  X  & --- & --- \\
Increased resolution   & --- &  X  &  X  & --- & --- & --- \\
Uses life information  & --- &  X  &  X  &  X  &  X  & --- \\
Uses early resets      & --- &  X  & --- &  X  & --- & --- \\
Separate eval episodes &  X  &  X  &  X  &  X  &  X  & --- \\
\bottomrule

\end{mytabular}
\vspace{-0.5ex}
\caption{Evaluation protocols for the Atari 100k benchmark. Computational resources are converted to A100 GPU days. EfficientMuZero\citep{ye2021effzero} achieves the highest scores but changed the environment configuration from the standard\citep{kaiser2019simple}. IRIS uses a separate hyperparameter for its exploration strength on Freeway.}
\label{tab:atari100k_setting}
\vspace*{-1ex}
\end{table}

\clearpage
\subsection*{Proprioceptive control learning curves}
\begin{figure}[h!]
\centering
\includegraphics[width=1\linewidth]{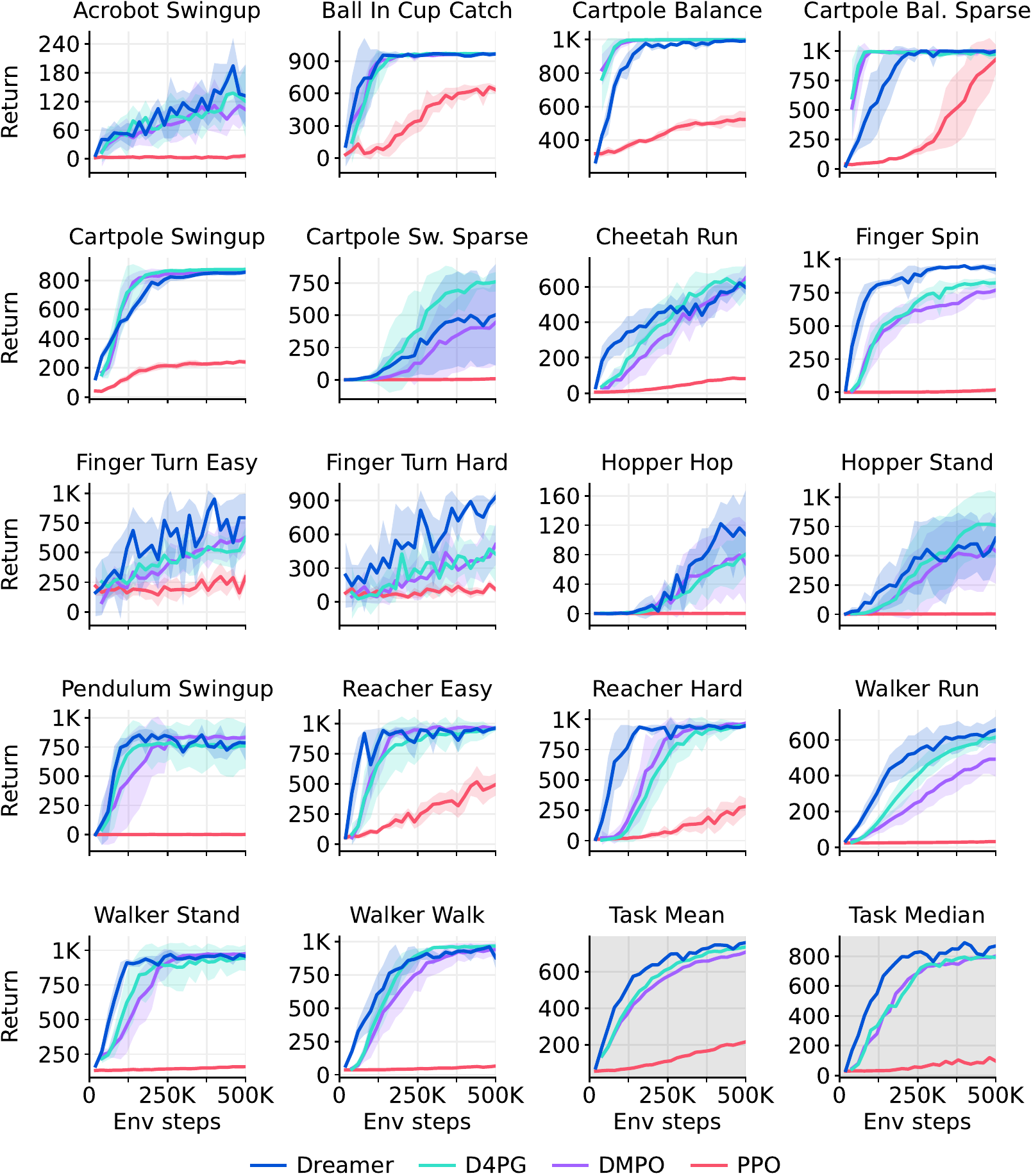}
\caption{DeepMind Control Suite learning curves under proprioceptive inputs.}
\label{fig:dmcp}
\end{figure}

\clearpage
\subsection*{Proprioceptive control scores}
\begin{table*}[h!]
\centering
\begin{mytabular}{
  colspec = {| L{12em} | C{4em} C{4em} C{4em} C{4em} C{4em} |},
  row{1} = {font=\bfseries},
}

\toprule
Task & PPO & DDPG & DMPO & D4PG & Dreamer \\
\midrule
Environment steps & 500K & 500K & 500K & 500K & 500K \\
\midrule

Acrobot Swingup & \o\o6 & 100 & 103 & 124 & \textbf{134} \\
Ball In Cup Catch & 632 & 917 & \textbf{968} & \textbf{968} & \textbf{962} \\
Cartpole Balance & 523 & \textbf{997} & \textbf{999} & \textbf{999} & \textbf{990} \\
Cartpole Balance Sparse & 930 & \textbf{992} & \textbf{999} & \textbf{974} & \textbf{990} \\
Cartpole Swingup & 240 & \textbf{864} & \textbf{860} & \textbf{875} & \textbf{852} \\
Cartpole Swingup Sparse & \o\o7 & 703 & 438 & \textbf{752} & 491 \\
Cheetah Run & \o82 & 596 & \textbf{650} & \textbf{624} & 614 \\
Finger Spin & \o18 & 775 & 769 & 823 & \textbf{931} \\
Finger Turn Easy & 281 & 499 & 620 & 612 & \textbf{793} \\
Finger Turn Hard & 106 & 313 & 495 & 421 & \textbf{889} \\
Hopper Hop & \o\o0 & \o36 & \o68 & \o80 & \textbf{113} \\
Hopper Stand & \o\o3 & 484 & 549 & \textbf{762} & 576 \\
Pendulum Swingup & \o\o1 & 767 & \textbf{834} & 759 & 788 \\
Reacher Easy & 494 & \textbf{934} & \textbf{961} & \textbf{960} & \textbf{954} \\
Reacher Hard & 288 & \textbf{949} & \textbf{968} & \textbf{937} & \textbf{938} \\
Walker Run & \o31 & 561 & 493 & 616 & \textbf{649} \\
Walker Stand & 159 & \textbf{965} & \textbf{975} & \textbf{947} & \textbf{964} \\
Walker Walk & \o64 & \textbf{952} & \textbf{942} & \textbf{969} & \textbf{936} \\
\midrule
Task mean & \o94 & 771 & 801 & 792 & \textbf{871} \\
Task median & 215 & 689 & 705 & \textbf{733} & \textbf{754} \\

\bottomrule
\end{mytabular}
\caption{DeepMind Control Suite scores under proprioceptive inputs.}
\label{tab:dmc_proprio}
\end{table*}

\clearpage
\subsection*{Visual control learning curves}
\begin{figure}[h!]
\centering
\vspace*{-1ex}
\includegraphics[width=1\linewidth]{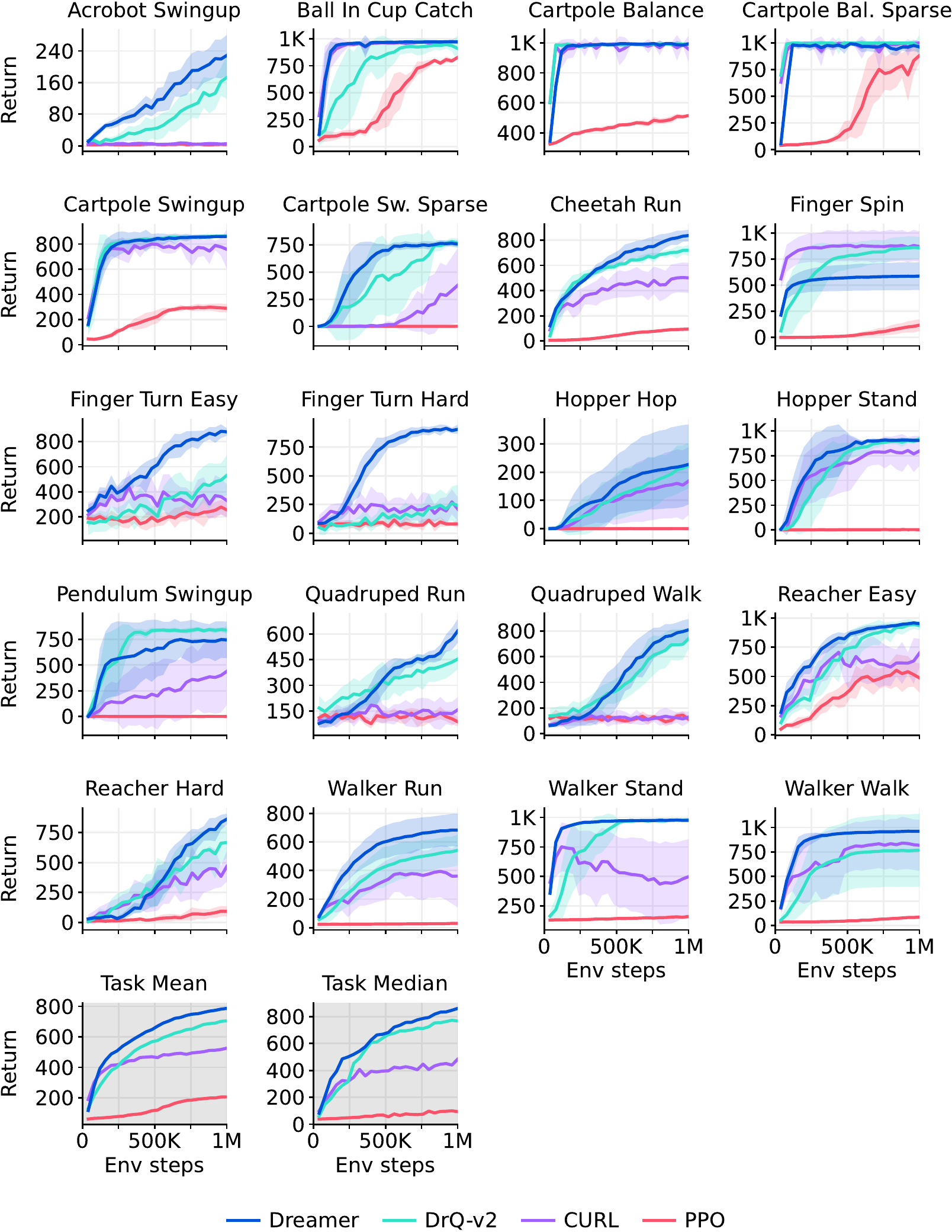}
\caption{DeepMind Control Suite learning curves under visual inputs.}
\label{fig:dmcv}
\vspace*{-2ex}
\end{figure}

\clearpage
\subsection*{Visual control scores}
\begin{table*}[h!]
\centering
\begin{mytabular}{
  colspec = {| L{12em} | C{4em} C{4em} C{4em} C{4em} C{4em} |},
  row{1} = {font=\bfseries},
}

\toprule
Task & PPO & SAC & CURL & DrQ-v2 & Dreamer \\
\midrule
Environment steps & 1M & 1M & 1M & 1M & 1M \\
\midrule

Acrobot Swingup & \o\o3 & \o\o4 & \o\o4 & 166 & \textbf{229} \\
Ball In Cup Catch & 829 & 176 & \textbf{970} & \textbf{928} & \textbf{972} \\
Cartpole Balance & 516 & 937 & \textbf{980} & \textbf{992} & \textbf{993} \\
Cartpole Balance Sparse & 881 & \textbf{956} & \textbf{999} & \textbf{987} & \textbf{964} \\
Cartpole Swingup & 290 & 706 & 771 & \textbf{863} & \textbf{861} \\
Cartpole Swingup Sparse & \o\o1 & 149 & 373 & \textbf{773} & \textbf{759} \\
Cheetah Run & \o95 & \o20 & 502 & 716 & \textbf{836} \\
Finger Spin & 118 & 291 & \textbf{880} & \textbf{862} & 589 \\
Finger Turn Easy & 253 & 200 & 340 & 525 & \textbf{878} \\
Finger Turn Hard & \o79 & \o94 & 231 & 247 & \textbf{904} \\
Hopper Hop & \o\o0 & \o\o0 & 164 & \textbf{221} & \textbf{227} \\
Hopper Stand & \o\o4 & \o\o5 & 777 & \textbf{903} & \textbf{903} \\
Pendulum Swingup & \o\o1 & 592 & 413 & \textbf{843} & 744 \\
Quadruped Run & \o88 & \o54 & 149 & 450 & \textbf{617} \\
Quadruped Walk & 112 & \o49 & 121 & 726 & \textbf{811} \\
Reacher Easy & 487 & \o67 & 689 & \textbf{944} & \textbf{951} \\
Reacher Hard & \o94 & \o\o7 & 472 & 670 & \textbf{862} \\
Walker Run & \o30 & \o27 & 360 & 539 & \textbf{684} \\
Walker Stand & 161 & 143 & 486 & \textbf{978} & \textbf{976} \\
Walker Walk & \o87 & \o40 & 822 & 768 & \textbf{961} \\

\midrule
Task mean & \o94 & \o81 & 479 & 770 & \textbf{861} \\
Task median & 206 & 226 & 525 & 705 & \textbf{786} \\
\bottomrule

\end{mytabular}
\caption{DeepMind Control Suite scores under visual inputs.}
\label{tab:dmc_vision}
\end{table*}

\clearpage
\subsection*{BSuite performance spectrum}
\begin{figure}[h!]
\centering
\includegraphics[width=0.8\linewidth,trim={5.3cm .5cm 2cm .5cm},clip]{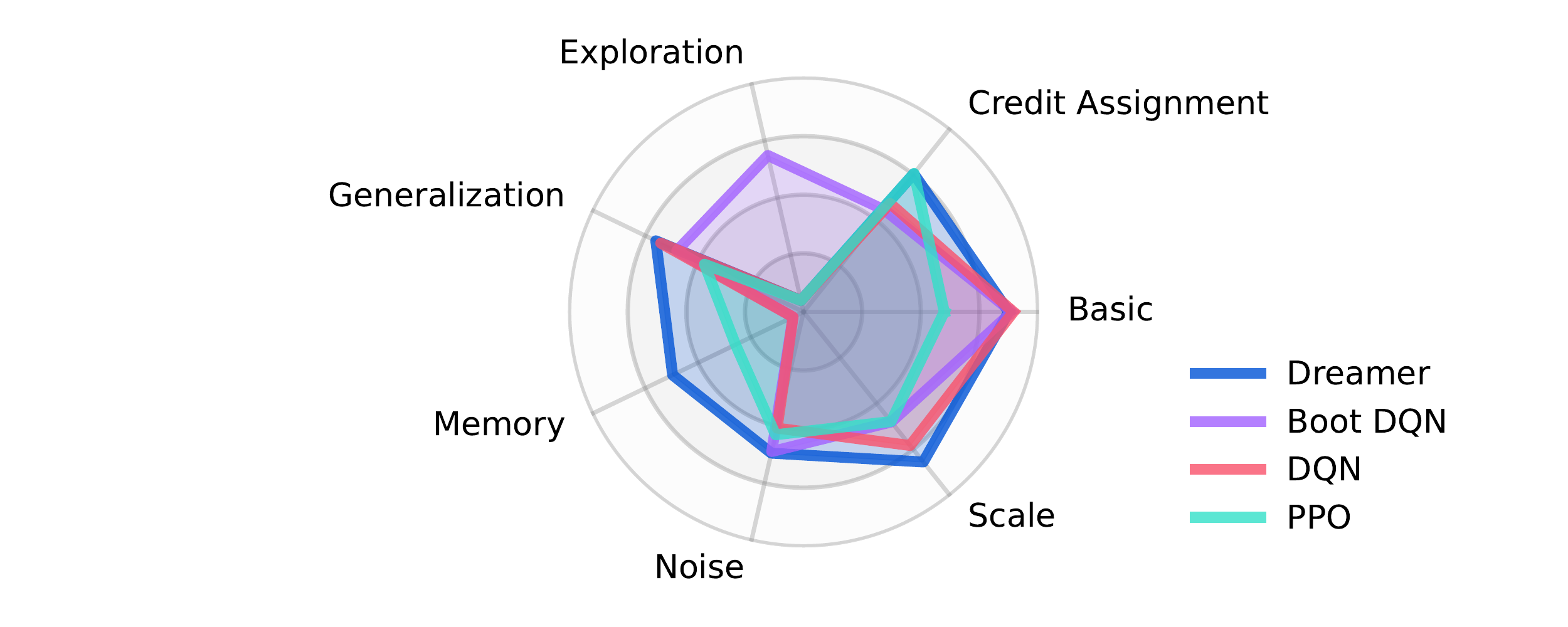}%
\caption{BSuite scores visualized by category\citep{osband2019bsuite}. Dreamer exceeds previous methods in the categories scale and memory. The scale category measure robustness to reward scales.}
\label{fig:bsuite}
\end{figure}

\clearpage
\subsection*{BSuite scores}
\begin{table}[h!]
\centering
\begin{mytabular}{
  colspec = {| L{10em} | C{3.8em} C{3.8em} C{3.8em} C{3.8em} C{3.8em} C{3.8em} |},
  row{1} = {font=\bfseries},
}
\toprule
Task & Random & PPO & \clap{AC-RNN} & DQN & \clap{Boot DQN} & Dreamer \\
\midrule
Bandit & 0.00 & 0.38 & \textbf{1.00} & 0.93 & \textbf{0.98} & \textbf{0.96} \\
Bandit Noise & 0.00 & 0.61 & 0.63 & 0.71 & \textbf{0.80} & 0.75 \\
Bandit Scale & 0.00 & 0.39 & 0.60 & 0.74 & \textbf{0.83} & 0.78 \\
Cartpole & 0.04 & 0.84 & 0.40 & 0.85 & 0.69 & \textbf{0.93} \\
Cartpole Noise & 0.04 & 0.77 & 0.20 & 0.82 & 0.69 & \textbf{0.93} \\
Cartpole Scale & 0.04 & 0.83 & 0.12 & 0.72 & 0.65 & \textbf{0.92} \\
Cartpole Swingup & 0.00 & 0.00 & 0.00 & 0.00 & \textbf{0.15} & 0.03 \\
Catch & 0.00 & 0.91 & 0.87 & 0.92 & \textbf{0.99} & \textbf{0.96} \\
Catch Noise & 0.00 & 0.54 & 0.27 & 0.58 & \textbf{0.68} & 0.53 \\
Catch Scale & 0.00 & \textbf{0.90} & 0.17 & 0.85 & 0.65 & \textbf{0.94} \\
Deep Sea & 0.00 & 0.00 & 0.00 & 0.00 & \textbf{1.00} & 0.00 \\
Deep Sea Stochastic & 0.00 & 0.00 & 0.00 & 0.00 & \textbf{0.90} & 0.00 \\
Discounting Chain & 0.20 & 0.24 & \textbf{0.39} & 0.25 & 0.22 & \textbf{0.40} \\
Memory Len & 0.00 & 0.17 & \textbf{0.70} & 0.04 & 0.04 & 0.65 \\
Memory Size & 0.00 & 0.47 & 0.29 & 0.00 & 0.00 & \textbf{0.59} \\
Mnist & 0.05 & 0.77 & 0.56 & \textbf{0.85} & \textbf{0.85} & 0.61 \\
Mnist Noise & 0.05 & \textbf{0.41} & 0.22 & 0.38 & 0.34 & 0.34 \\
Mnist Scale & 0.05 & \textbf{0.76} & 0.09 & 0.49 & 0.31 & 0.55 \\
Mountain Car & 0.10 & 0.10 & 0.10 & \textbf{0.93} & \textbf{0.93} & \textbf{0.92} \\
Mountain Car Noise & 0.10 & 0.10 & 0.10 & \textbf{0.89} & 0.82 & \textbf{0.87} \\
Mountain Car Scale & 0.10 & 0.10 & 0.10 & 0.85 & 0.56 & \textbf{0.90} \\
Umbrella Distract & 0.00 & \textbf{1.00} & 0.09 & 0.30 & 0.26 & 0.74 \\
Umbrella Length & 0.00 & \textbf{0.87} & 0.43 & 0.39 & 0.39 & 0.78 \\
\midrule
Basic & 0.04 & 0.60 & 0.58 & \textbf{0.90} & \textbf{0.89} & \textbf{0.88} \\
Credit assignment & 0.03 & \textbf{0.76} & 0.37 & 0.59 & 0.56 & \textbf{0.75} \\
Exploration & 0.00 & 0.00 & 0.00 & 0.00 & \textbf{0.68} & 0.01 \\
Generalization & 0.06 & 0.47 & 0.19 & \textbf{0.68} & 0.60 & \textbf{0.70} \\
Memory & 0.00 & 0.32 & 0.49 & 0.02 & 0.02 & \textbf{0.62} \\
Noise & 0.02 & 0.54 & 0.24 & 0.51 & \textbf{0.61} & \textbf{0.62} \\
Scale & 0.04 & 0.60 & 0.22 & 0.73 & 0.60 & \textbf{0.82} \\
\midrule
Task mean (\%) & \o3 & 49 & 32 & 54 & 60 & \textbf{66} \\
Category mean (\%) & \o3 & 47 & 30 & 49 & 57 & \textbf{63} \\
\bottomrule

\end{mytabular}
\caption{BSuite scores for each task averaged over environment configurations, as well as aggregated performance by category and over all tasks.}
\label{tab:bsuite}
\end{table}

\clearpage
\subsection*{Robustness ablations}
\begin{figure}[h!]
\centering
\includegraphics[width=\linewidth]{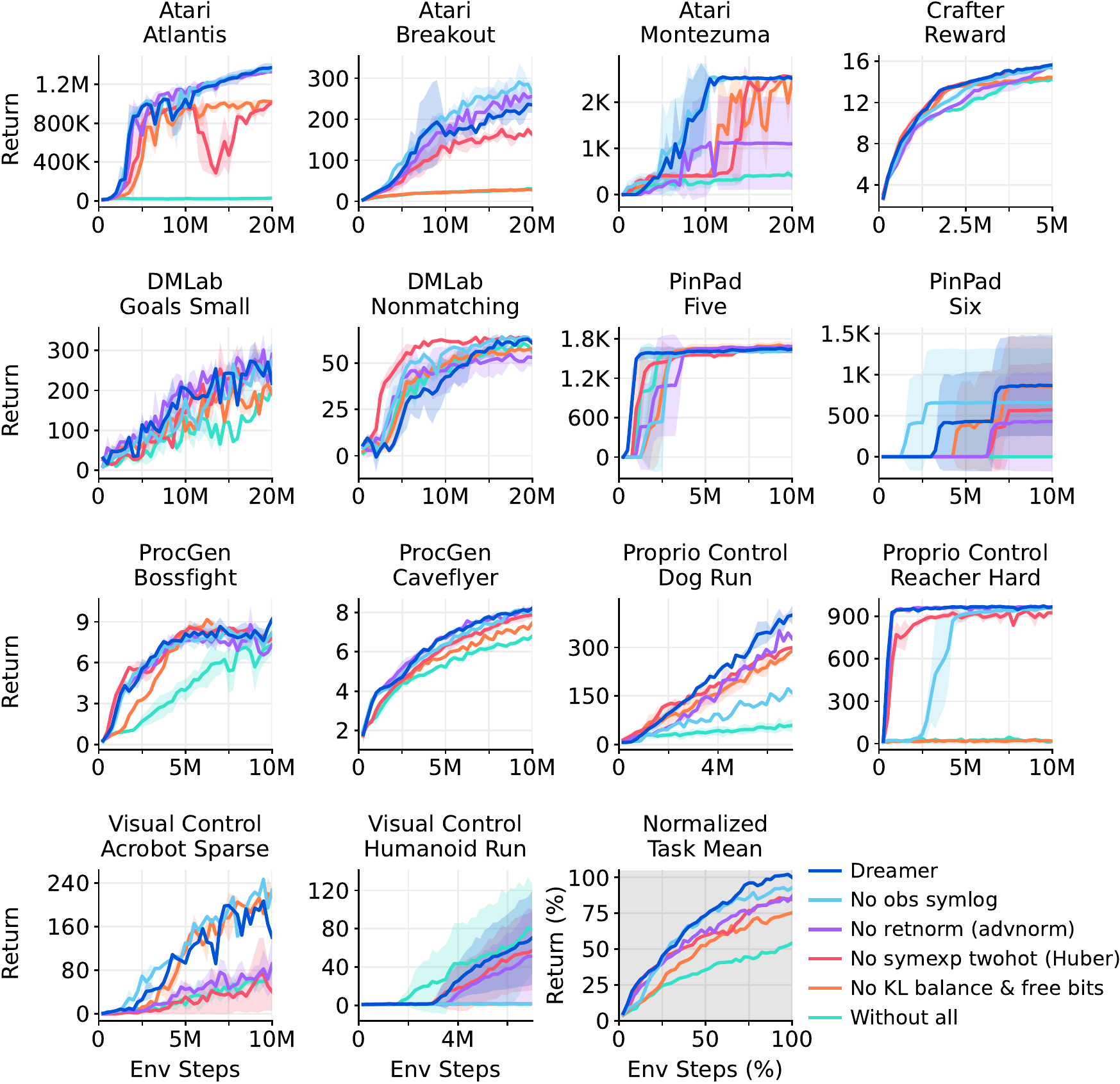}
\caption{Individual learning curves for the robustness ablation experiment. All robustness techniques contribute to the overall performance of Dreamer, although each individual technique may only improve the performance on a subset of the tasks.}
\label{fig:robustness}
\end{figure}

\clearpage
\subsection*{Learning signal ablations}
\begin{figure}[h!]
\centering
\includegraphics[width=\linewidth]{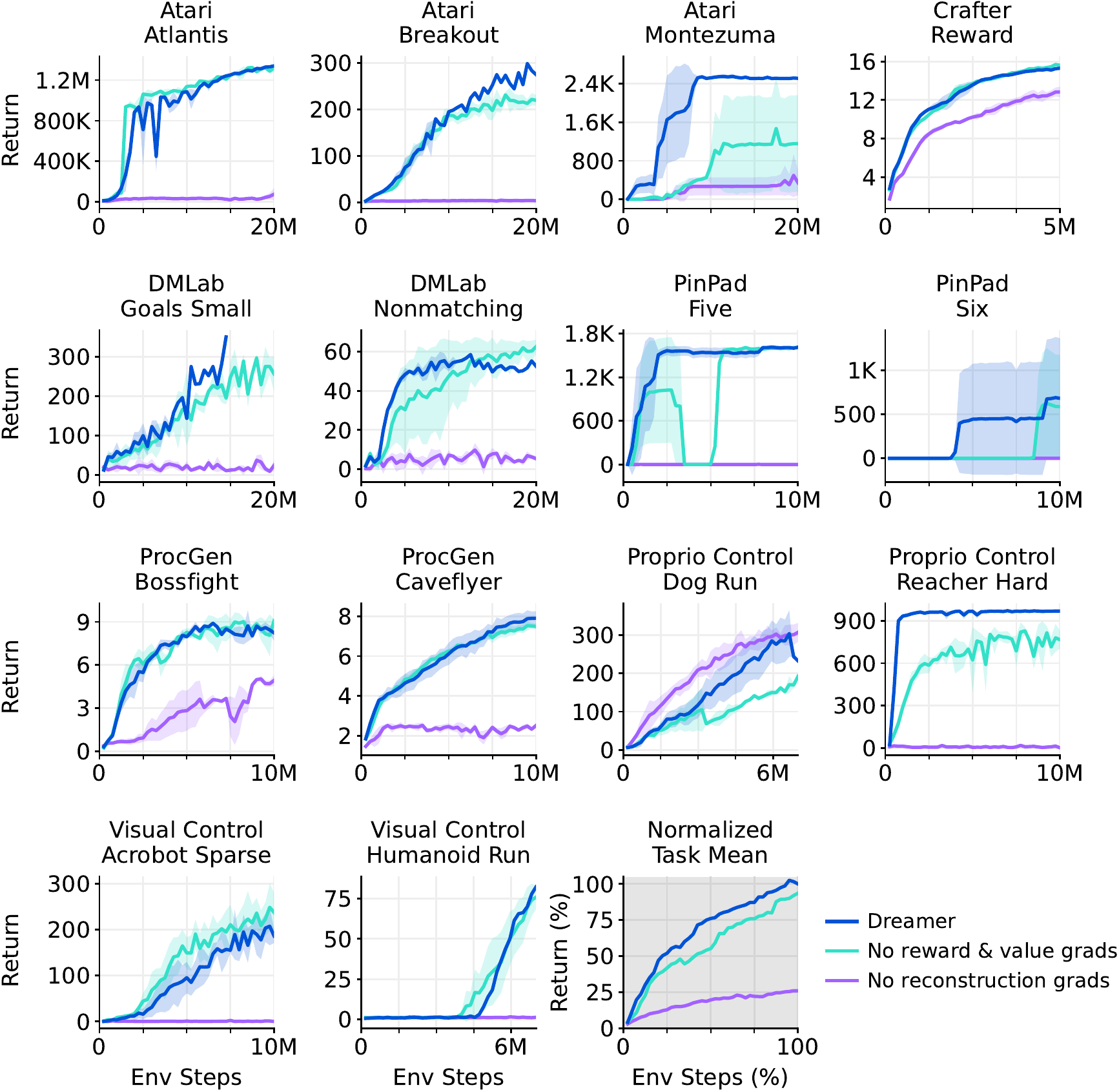}
\caption{Individual learning curves for the learning signal ablation experiment. Dreamer relies predominantly on the undersupervised reconstruction objective of its world model and additional reward and value gradients further improve performance on a subset of tasks.}
\label{fig:signal}
\end{figure}

\end{document}